\documentclass[twocolumn, 10pt, A4paper]{article}

\usepackage{geometry}
\usepackage{newtxtext}
\usepackage{newtxmath} 
\usepackage[justification=centering]{caption}

\usepackage{enumitem}

\usepackage{natbib}

\usepackage{amsmath}
\usepackage{epsfig}
\usepackage{graphicx}
\usepackage{vmargin}
\usepackage{url}
\usepackage{tabularx}
\usepackage{boxedminipage}

\usepackage{lastpage} 
\usepackage{fancyhdr}
\usepackage{soul} 
\usepackage{stfloats}

\usepackage[linesnumbered]{algorithm2e}   
\SetAlCapSkip{1em}          
\RestyleAlgo{ruled}

\usepackage{subcaption}
\usepackage[capitalize]{cleveref}
\crefname{algocf}{Alg.}{Algs.}
\Crefname{algocf}{Algorithm}{Algorithms}
\crefname{equation}{Eq.}{Eqs.}
\Crefname{equation}{Equation}{Equations}
\crefname{figure}{Fig.}{Figs.}
\Crefname{figure}{Figure}{Figures}
\crefname{method}{Meth.}{Meths.}
\Crefname{method}{Method}{Methods}
\crefname{section}{Sec.}{Secs.}
\Crefname{section}{Section}{Sections}
\crefname{table}{Tab.}{Tabs.}
\Crefname{table}{Table}{Tables}

\usepackage{booktabs}
\usepackage{multirow}

\usepackage{microtype}

\newcommand{\captionfonts}{\fontsize{9}{9}\it}
\makeatletter  
\long\def\@makecaption#1#2{ %
  \vskip\abovecaptionskip
  \sbox\@tempboxa{{\captionfonts #1: #2}}%
   \ifdim \wd\@tempboxa >\hsize
	{\centering\captionfonts #1: #2\par}
   \else
    \hbox to\hsize{\hfil\box\@tempboxa\hfil}%
   \fi
  \vskip\belowcaptionskip}
\makeatother   
 
\vfuzz=\hfuzz

\newlength{\defbaselineskip}
\newcommand{\setlinespacing}[1]%
           {\setlength{\baselineskip}{#1 \defbaselineskip}}

\makeatletter
\newsavebox\saved@arstrutbox
\newcommand*{\setarstrut}[1]{%
	\noalign{%
		\begingroup
		\global\setbox\saved@arstrutbox\copy\@arstrutbox
		\global\setbox\@arstrutbox\hbox{%
			\vrule \@height #1
			\@depth  0cm
			\@width\z@
		}%
		\endgroup
	}%
}
\newcommand*{\restorearstrut}{%
	\noalign{%
		\global\setbox\@arstrutbox\copy\saved@arstrutbox
	}%
}
\makeatother

\makeatletter

\renewcommand\section{\@startsection{section}{1}{\z@}{6pt}{3pt}{\normalfont\large\bfseries}}
\renewcommand\subsection{\@startsection{subsection}{1}{\z@}{6pt}{3pt}{\normalfont\normalsize\bfseries}}

\renewcommand\subsubsection{\@startsection{subsubsection}{1}{\z@}{6pt}{3pt}{\normalfont\normalsize\itshape}}

\makeatother

\newcommand{\authorfont}{\fontsize{11}{14}\selectfont}
\newcommand{\titlefont}{\fontsize{12}{14}\selectfont \bf}

\begin{document}

\fancypagestyle{empty}{
  \vspace{5pt}
  \fancyhf{}
  \renewcommand{\headrulewidth}{2.25pt}
  \renewcommand{\headsep}{13pt}

  \fancyhead[L]{%
    \begin{tabular}[b]{@{}l@{}}
      \includegraphics[height=1.8cm, trim={0 0.1cm 0 0}, clip]{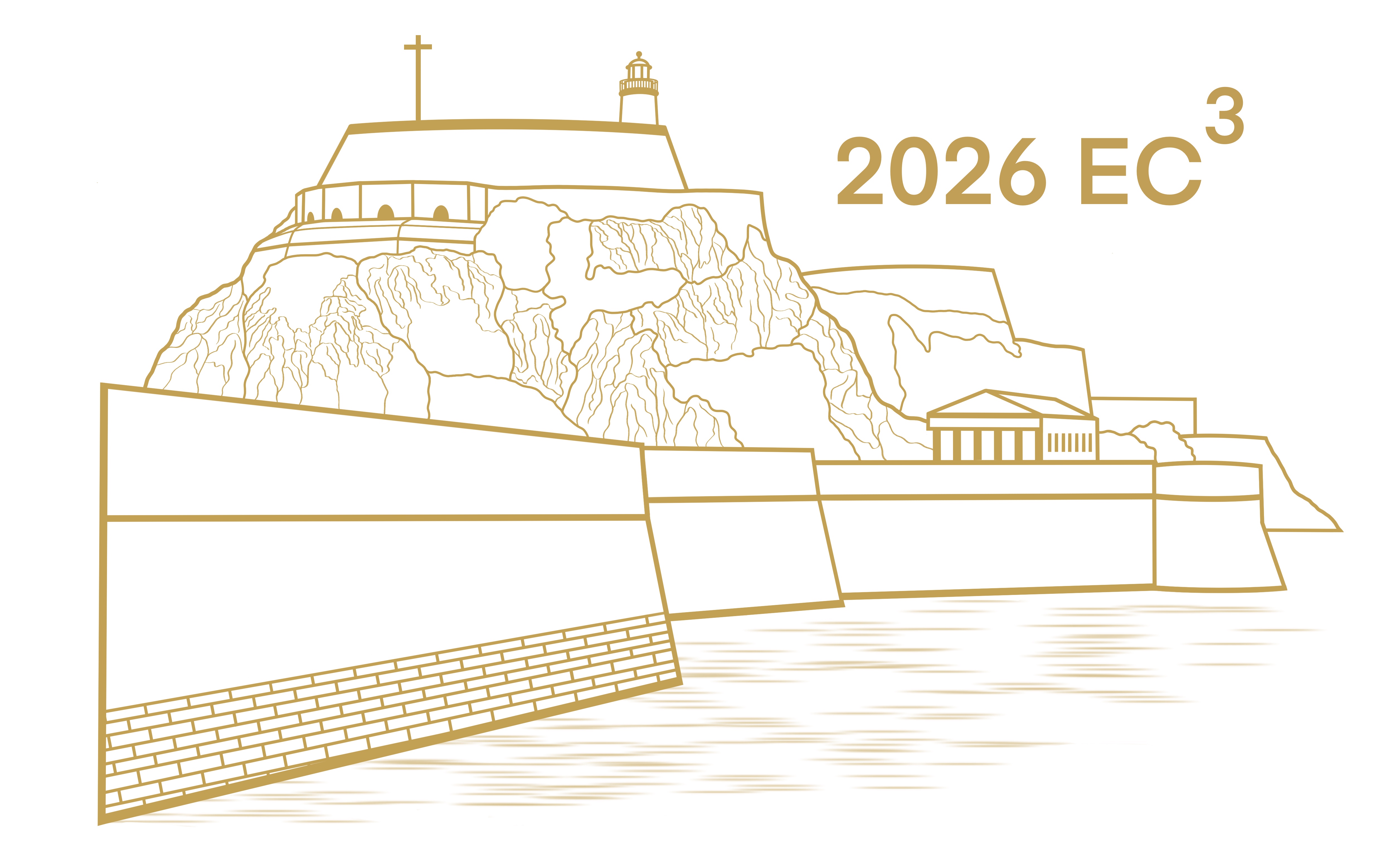}
    \end{tabular}
  }

  \fancyhead[C]{%
    \begin{tabular}[b]{c}
      2026 European Conference on Computing in Construction \\
      Corfu, Greece \\
      July 12–15, 2026
    \end{tabular}
  }

  \fancyhead[R]{%
    \begin{tabular}[b]{@{}r@{}}
      \includegraphics[height=1.45cm, trim={0 0.1cm 0 0}, clip]{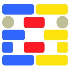}
    \end{tabular}
  }
}

\bibliographystyle{apalikenonitalic}
\date{}
\title{\vspace{-9mm} \titlefont BIMStruct3D: A Fully Automated Hybrid Learning Scan-to-BIM Pipeline with Integrated Topology Refinement \vspace{-5.5mm}}
\author{
\authorfont{~}\\
\authorfont{Mahdi Chamseddine$^{1,2}$, Fabian Kaufmann$^2$, Marius Schellen$^2$, Christian Glock$^2$,}\\
\authorfont{Didier Stricker$^{1,2}$, and Jason Rambach$^{1}$}\\
\authorfont{~}\\
\authorfont{$^1$German Research Center for Artificial Intelligence (DFKI), Kaiserslautern, Germany}\\
\authorfont{$^2$RPTU Kaiserslautern-Landau, Kaiserslautern, Germany}\\
\authorfont{\textit{\small\color{gray!70!black} Accepted for presentation. This pre-print includes supplementary material that is not part of the official conference proceedings.}}\\
\authorfont{~}\\
\vspace{-15mm}
}
\maketitle
\thispagestyle{empty}

\setul{2pt}{0.25pt} 
\section{Abstract}
Automatic generation of Building Information Models (BIM) from building scans is a key challenge in architecture and construction. We present a modular pipeline for generating IFC-compliant BIM from 3D point clouds. The hybrid approach combines learning-based semantic segmentation with topology-aware geometric reconstruction to model structural elements accurately. We propose vIoU, adapting voxel-based overlap evaluation to Scan-to-BIM by enabling holistic, instance-matching-free comparison of reconstructed and ground-truth models. We release the German Hospital dataset (DeKH), including high-resolution point clouds, ground truth BIMs, and semantic annotations. Experiments on DeKH and CV4AEC datasets show significant improvements over a RANSAC-based baseline, demonstrating robustness and scalability.

\addtocounter{section}{1}
\section{Introduction} \label{sec:Introduction}

Capturing spatial data of existing built environments can be efficiently achieved using mobile~\citep{kanayama2025tof} and terrestrial scanning systems~\citep{chang2017matterport3d,Armeni.2017}. To leverage such data in construction~\citep{wu2021application} and facility management~\citep{xu20213d}, it must be converted into semantic models in accordance with the Building Information Modelling (BIM) methodology. Given the volume of data and the complex steps required to transform raw 3D point clouds into BIM models, full automation of this process is highly desirable. However, this transformation remains a challenge.

\begin{figure}[t]
    \renewcommand\thesubfigure{\Roman{subfigure}}
    \centering
    \begin{subfigure}[t]{0.24\linewidth}
        \includegraphics[width=\linewidth]{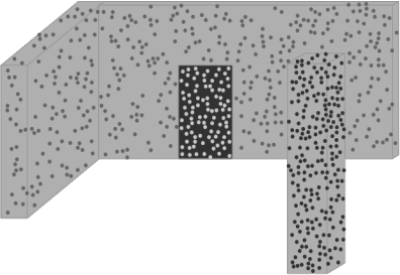}
        \caption{\scriptsize Input point cloud}
    \end{subfigure}
    \begin{subfigure}[t]{0.24\linewidth}
        \includegraphics[width=\linewidth]{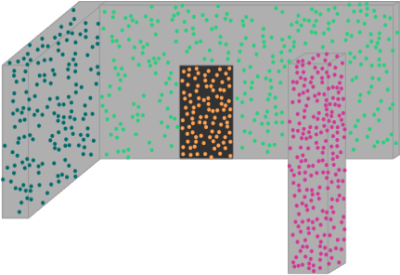}
        \caption{\scriptsize Semantic segmentation}
    \end{subfigure}
    \begin{subfigure}[t]{0.24\linewidth}
        \includegraphics[width=\linewidth]{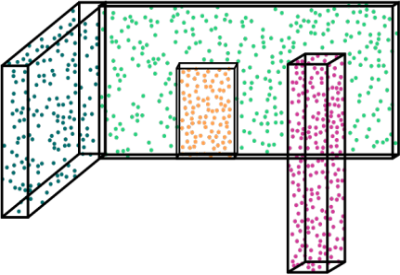}
        \caption{\scriptsize Geometric reconstruction}
    \end{subfigure}
    \begin{subfigure}[t]{0.24\linewidth}
        \includegraphics[width=\linewidth]{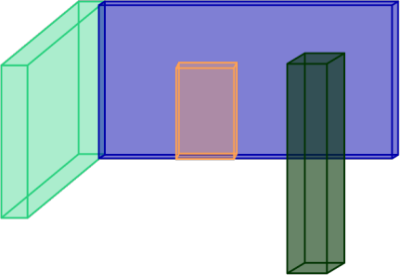}
        \caption{\scriptsize BIM model}
    \end{subfigure}
    \vspace{0.5em}
    \caption{BIMStruct3D is a hybrid pipeline for generating IFC BIM models from 3D point cloud scans}
    \label{fig:teaser}
    \vspace{-1em}
\end{figure}

Despite the importance of the task, scan-to-BIM is rarely treated as a comprehensive, end-to-end problem with standardized benchmarks; however, some effort is being made through initiatives like the CV4AEC challenges~\citep{armeni2024cv4aec}. In this work, we introduce BIMStruct3D, one of the first comprehensive frameworks for automating the conversion of noisy and incomplete 3D point clouds into IFC-compatible BIM models. Our BIMStruct3D pipeline employs a hybrid approach that integrates both learning-based and classical point cloud processing techniques. It includes modules for 3D point cloud segmentation, geometric reconstruction and primitive extraction, as well as model post-processing, as illustrated in~\Cref{fig:teaser}.

BIMStruct3D is designed to process data from multi-storey buildings such as offices, hospitals, and residential complexes, and generate BIM representations of architectural elements including walls, doors, and columns. Accompanying our framework is \texttt{pystruct3d}, a modular library tailored for processing 3D point cloud data of buildings.

To support reproducibility and facilitate benchmarking in the scan-to-BIM domain, we provide a comprehensive evaluation of our method at both modular and end-to-end levels. This includes tests on publicly available datasets and newly acquired data, featuring high-precision, handcrafted ground truth BIMs. We also propose volumetric Intersection over Union (vIoU), which adapts voxel-based IoU evaluation to the Scan-to-BIM domain. Unlike per-object voxel IoU used in 3D shape reconstruction, vIoU operates at the class level without instance matching, making it robust to the fragmented and variably segmented wall representations common in BIM reconstruction.

In real-world scenarios, building point clouds are often incomplete due to factors such as the limitations of scanning technology, human error and environmental conditions like lighting, occlusion, and material reflectivity (e.g. glass surfaces). BIMStruct3D is built to handle such noisy and incomplete data to create usable BIM representations.

In summary, our contributions are as follows:
\begin{itemize}
    \item BIMStruct3D, a fully automated hybrid pipeline generating IFC BIM models from 3D point clouds using semantic segmentation and geometric reconstruction.
    \item Specialized algorithms for primitive extraction and BIM model generation.
    \item vIoU, an adaptation of voxel-based IoU to Scan-to-BIM evaluation, enabling class-level comparison without instance-level matching.
    \item \texttt{pystruct3d}\footnote{\url{https://github.com/humantecheu/pystruct3d}}, an open-source library for processing 3D building point clouds, released with DeKH\footnote{\url{https://huggingface.co/datasets/RPTU-FGMB/DeKH}}, a new dataset including high-resolution scans, semantic annotations, and handcrafted ground truth BIMs.
\end{itemize}

\section{Related Work} \label{sec:RelatedWork}

\subsection{Point Cloud Semantic Segmentation}

Semantic segmentation of point clouds is a fundamental computer vision task, currently dominated by deep neural networks. Early approaches include projection-based~\citep{chen2017multi}, voxel-based~\citep{Graham.2018}, and point-based networks~\citep{Qi.2017}, each trading off between computational cost, geometric fidelity, and flexibility. More recent developments have incorporated self-attention mechanisms, demonstrating strong performance in large-scale 3D scene understanding~\citep{wang2019dynamic}. Point Transformer by~\cite{Zhao.2021} advanced point-based segmentation by integrating local self-attention~\citep{Vaswani.2017}, vector attention~\citep{zhao2020exploring}, and appropriate positional encodings. Subsequent iterations~\citep{wu2024point} introduce architectural improvements leading to better segmentation accuracy and model efficiency.

\subsection{Scan-to-BIM Approaches}

To align with the scope of this work, we focus on research targeting fully automated (end-to-end) scan-to-BIM conversion from point clouds. \cite{Gourguechon.2022} and~\cite{Bassier.2020} distinguish between room-based and wall-based (or structural component-based) reconstruction approaches. The former identifies rooms and then reconstructs surrounding elements, while the latter directly targets structural components such as walls. Unfortunately, to this day, there are no open-source implementations of complete scan-to-BIM pipelines or benchmark datasets for evaluation.

\subsubsection*{Room-based reconstruction}

Room-based methods segment indoor spaces and reconstruct surrounding elements. Representative approaches address plane fitting and wall estimation~\citep{Ochmann.2016,Ochmann.2019}, 2D region growing with slab reconstruction~\citep{Macher.2017}, comprehensive element reconstruction including windows and fixtures~\citep{Xiong.2023}, hybrid IFC wall reconstruction with topology refinement~\citep{Bassier.2018}, and learning-based segmentation followed by space decomposition~\citep{Tang.2022,Hu.2020}. While these methods offer good indoor segmentation, \cite{Bassier.2020} note their limitations in modelling non-room structures and handling clutter. Thus, component-based approaches targeting structural elements have gained traction.

\subsubsection*{Structural component reconstruction}

Component-based reconstruction identifies and models structural elements like walls and columns through clustering, semantic segmentation, and primitive fitting. \cite{Bassier.2020} focused on extracting standard wall objects with topological rules for merging, while follow-up work~\citep{Bassier.2020b} compared 2D and 3D reconstruction methods. Other approaches combine 2D and 3D processing with point cloud upsampling~\citep{Gankhuyag.2021}, use RANSAC and convex hull-based boundary extraction~\citep{Thomson.2015}, reconstruct walls and spaces from enclosed rooms~\citep{Anagnostopoulos.2016}, or segment concrete components via concavity/convexity criteria~\citep{Son.2017}. The closest work to ours is by \cite{Kim.2021}, which uses synthetic data to train a segmentation model, followed by instance clustering and planar patch segmentation. Element relationships are modelled with graph networks to infer missing links and extend incomplete elements.

Our approach differs in that we do not rely on synthetic data and instead address several common limitations, including limited generalization and reliance on manual parameter tuning. We integrate semantic segmentation, geometric reconstruction, and topological reasoning into a unified scan-to-BIM pipeline.

\begin{figure*}[htbp]
    \centering
    \includegraphics[width=0.9\textwidth]{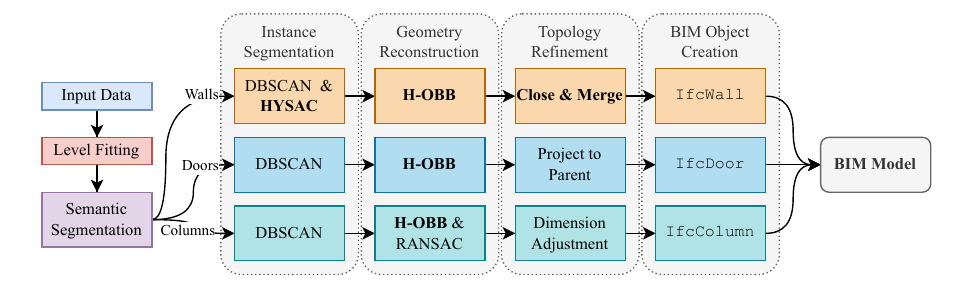}
    \caption{Overview of the reconstruction pipeline showing the different reconstruction stages for walls, doors, and columns. In \textbf{bold} are our main contributed components.}
    \label{fig:overview}
    \vspace{-1em}
\end{figure*}
\section{Methodology} \label{sec:Methodology}

Our method, outlined in~\Cref{fig:overview}, consists of semantic and instance segmentation, geometry reconstruction, topology refinement, and BIM object creation from a 3D point cloud. IFC export is handled via \textit{IfcOpenShell}\footnote{\url{https://ifcopenshell.org/}}, generating a project--site--building--storey hierarchy and assigning reconstructed geometry as corresponding IFC entities (\texttt{IfcWall}, \texttt{IfcDoor}, \texttt{IfcColumn}). During the reconstruction process, we adopt the Manhattan world assumption~\citep{Coughlan.1999}, which holds for the majority of structural objects in typical building environments.

Initially, levels/storeys are identified by analysing vertical point density distributions, using surfaces like floors and ceilings to detect significant point concentrations~\citep{Bassier.2020b,Macher.2017}. Semantic segmentation using state-of-the-art deep learning models~\citep{wu2024point} serves as the basis for reconstructing walls, doors, and columns. The reconstruction pipeline follows a common structure: instance segmentation, geometry reconstruction, topology refinement, and IFC BIM object creation.

\begin{figure}[htbp]
    \centering
    \includegraphics[width=0.75\linewidth]{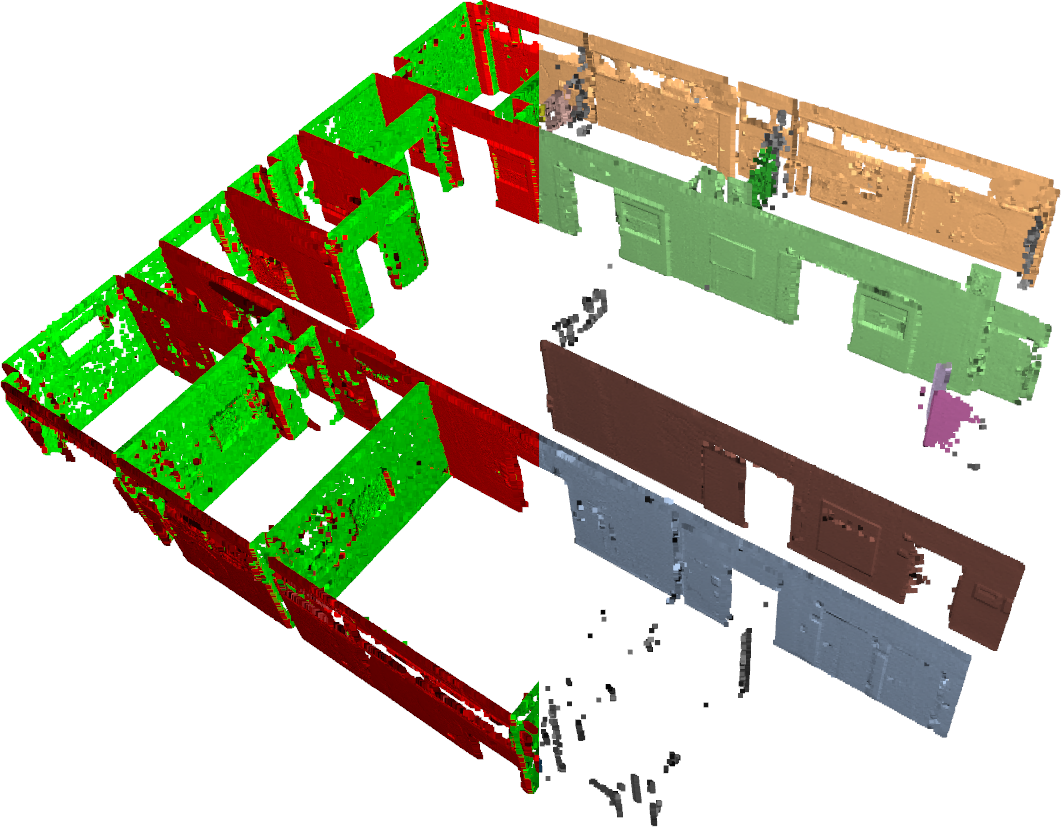}
    \caption{(Left) Wall direction filtering shown in red and green and (Right) DBSCAN clustering for one of the directions in multiple colours.}
    \label{fig:walls_dbscan}
    \vspace{-0.5em}
\end{figure}

\textbf{Wall} reconstruction is based on axis-aligned point grouping. DBSCAN clustering~\citep{Ester.1996} is applied per axis to form spatial clusters, discarding sparse noise as seen in~\Cref{fig:walls_dbscan}. HYSAC, our RANSAC~\citep{Schnabel.2007} based plane segmentation algorithm with optimized seed selection, is used to extract wall surfaces. These are enclosed using Horizontal Oriented Bounding Boxes (H-OBB), followed by topological refinements to ensure geometric correctness.

\textbf{Door} points are initially identified via semantic segmentation and then treated as child objects of walls, relying on the reconstructed wall geometry as a starting point. The process first identifies door points within the wall bounding boxes, which typically correspond to frames, linings, or closed door leaves captured during scanning. If such points are found, the bounding box is expanded to include additional door geometry, such as open door leaves, ensuring doors are correctly associated with their respective parent walls. Our two-step approach reduces errors caused by open doors adjacent to perpendicular walls.

\textbf{Column} points are clustered using DBSCAN~\citep{Ester.1996} and filtered by minimum point thresholds to reduce artifacts. Shape classification is achieved through curvature analysis based on eigenvalue decomposition from k-d tree neighbourhood searches. Curvature distributions are modelled with Maximum Likelihood Estimation to distinguish round (low standard deviation) from square or rectangular columns (high standard deviation). We then use RANSAC cylinder fitting for round columns and H-OBB fitting square/rectangular columns.

\subsection{Semantic Segmentation}

Semantic segmentation is essential for producing accurate IFC models, as it forms the foundation for all subsequent steps. High-quality segmentation ensures later geometry reconstruction stages are based on reliable object labels.

We use Point Transformer V3 by~\cite{wu2024point}, trained on a combined dataset from three public datasets: S3DIS~\citep{Armeni.2016}, Structured3D~\citep{zheng2020structured3d}, and ScanNet~\citep{dai2017scannet}. Training on a joint dataset prevents the model from over-fitting to the data and sensor used in one dataset. To further improve generalization, additional data augmentation techniques were used, such as removing color information to simulate scans from grayscale 3D sensors. This is particularly relevant because the CV4AEC dataset
used for the evaluation contains colored and grayscale point clouds. This segmentation pipeline ensures robust performance across diverse scan types and scene complexities, enabling reliable downstream reconstruction of BIM elements.

\subsection{Topology aware geometry reconstruction} 

The reconstruction pipeline leverages both state-of-the-art techniques and our novel contributions. The following sections describe our key innovations in wall reconstruction, oriented bounding box fitting, and topology correction.

\subsubsection{HYSAC wall reconstruction} \label{sec:hysac}

The RANSAC-based plane fitting methods~\citep{Schnabel.2007} are highly sensitive to parameter settings like the minimum number of inliers and distance thresholds. Improper parameter tuning often results in under-segmentation, where a single plane, instead of two, is incorrectly fitted to an entire wall instance.

To address this, we propose Hypothesis-based Sample Consensus (HYSAC), a modified RANSAC approach that improves seed point selection. Instead of sampling seed points randomly, HYSAC benefits from the point density distribution perpendicular to the primary wall orientation. Inspired by~\cite{Armeni.2016}, we compute a 1D density histogram along the axis orthogonal to the wall surface and select seed points from regions of highest density. Planes are fitted using Singular Value Decomposition (SVD), which provides a least-squares optimal plane fit robust to point cloud noise, and accepted based on a minimum inlier ratio. \Cref{algo:hysac} shows a pseudo code implementation of HYSAC.

\begin{algorithm}[t]
    \small
    \caption{HYSAC plane fitting pseudocode.} \label{algo:hysac}
    
    \DontPrintSemicolon
    \newcommand\KwFont[1]{\ttfamily\textbf{#1}}
    \newcommand\CommentFont[1]{\smaller\ttfamily\textcolor{blue}{#1}}
    
    \SetKwProg{Def}{\KwFont{def}}{\KwFont{:}}{\KwFont{end}}
    \SetKwFor{While}{\KwFont{while}}{\KwFont{:}}{\KwFont{end}}
    \SetKwIF{If}{ElseIf}{Else}{\KwFont{if}}{\KwFont{:}}{\KwFont{elif}}{\KwFont{else:}}{\KwFont{end}}
    \SetKwComment{Comment}{\# }{}
    \SetCommentSty{CommentFont}
    \SetKw{KwRet}{\KwFont{return}}
    \SetKw{Yield}{\KwFont{yield}}
    
    \SetKwInOut{Input}{\texttt{input}}
    \SetKwInOut{Output}{\texttt{output}}
    
    \SetKwFunction{Histogram}{histogram}
    \SetKwFunction{HistoSeeds}{histogram\_seeds}
    \SetKwFunction{RandomSample}{random\_sample}
    \SetKwFunction{FindPeaks}{find\_peaks}
    \SetKwFunction{FitPlaneSVD}{fit\_plane\_svd}
    \SetKwFunction{HysacPlane}{hysac\_plane}
    \SetKwFunction{NumOf}{number\_of}
    \SetKwFunction{Remove}{remove}

    \Input{wall\_points~\Comment{points of a wall cluster}} 
    \Input{n\_bins\hspace{20.7pt}\Comment{number of histogram bins}}
    \Input{n\_seeds\hspace{16.2pt}\Comment{number of seed points}}
    \Input{min\_points\hspace{3.5pt}\Comment{minimum points per plane}}
    \Output{planes\hspace{21.7pt}\Comment{equations of found planes}}
    \Output{inliers\hspace{4.3pt}\Comment{points belonging to the planes}}
    \BlankLine
    
    \Def{\HistoSeeds{wall\_points, n\_seeds}}{
        \Comment{generate the histogram perpendicular to the wall direction}
        $hist = \Histogram{cluster\_points, n\_bins}$\;
        $peak = \FindPeaks{hist}$\;
        \Comment{select random points from peaks}
        $seed\_points = \RandomSample{peak, n\_seeds}$\;
        \KwRet{seed\_points}
    }
    \BlankLine
    
    \Def{\HysacPlane{wall\_points, n\_seeds}}{
            $points = wall\_points$\;
            \Comment{loop until all planes are found}
            \While(){\NumOf{points} $\geq$ min\_points}{
                $seeds = \HistoSeeds{points, n\_seeds}$\;
                $plane, inliers = \FitPlaneSVD{seeds}$\; 
                \If{\NumOf{inliers} $\geq$ min\_points}{
                    \Comment{remove inliers of the plane from the wall points}
                    $points = \Remove{points, inliers}$\;
                    \Yield plane, inliers
            }
        }
    }
    \BlankLine

    $planes, inliers = \HysacPlane{wall\_points, n\_seeds}$\;
    \KwRet{planes, inliers}
\end{algorithm}

\begin{figure}[htbp]
    \centering
    \includegraphics[width=0.9\linewidth]{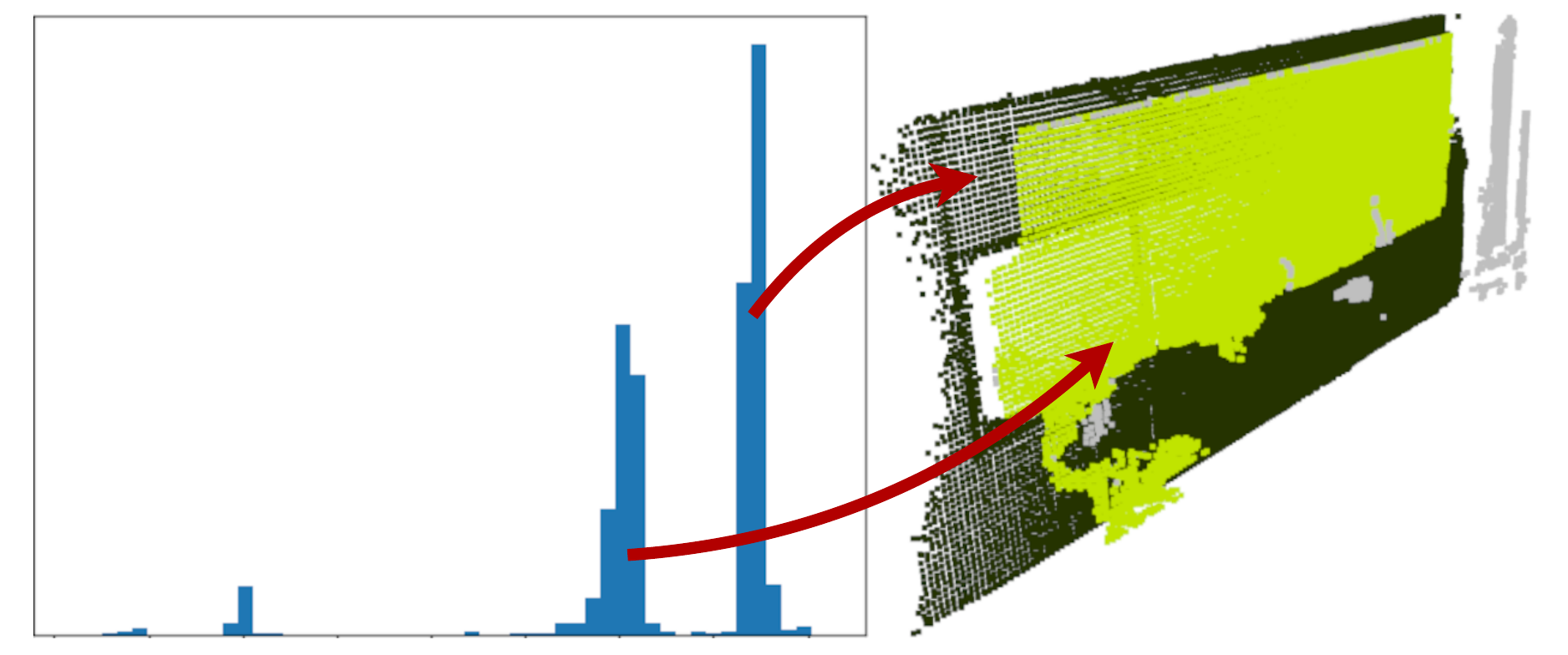}
    \caption{HYSAC plane fitting. Result of plane fitting (plane points coloured, outliers grey) with corresponding histogram of point distribution.}
    \label{fig:hysac}
    \vspace{-1em}
\end{figure}

As shown in~\Cref{fig:hysac}, HYSAC reliably detects multiple planes within a wall cluster, distinguishing between valid wall surfaces and outliers. The resulting inlier points are grouped and forwarded for further geometric processing.

\subsubsection{Horizontal Oriented Bounding Boxes Fitting} \label{sec:hobb}
To approximate the geometry of wall instances, we fit Horizontal Oriented Bounding Boxes (H-OBB) to the previously segmented plane clusters. Given that architectural structures typically conform to horizontal and vertical alignments, we restrict the bounding boxes to axis-aligned orientations within the XY-plane.

\begin{figure}[htbp]
    \renewcommand\thesubfigure{\Roman{subfigure}}
    \centering
    \begin{subfigure}[c]{0.2\linewidth}
        \includegraphics[width=\linewidth]{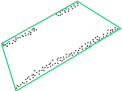}
        \caption{}
    \end{subfigure}
    \hspace{0.5em}
    \begin{subfigure}[c]{0.4\linewidth}
        \includegraphics[width=\linewidth]{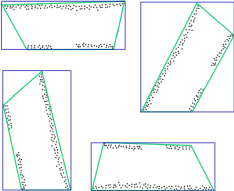}
        \caption{}
    \end{subfigure}
    \hspace{0.5em}
    \begin{subfigure}[c]{0.2\linewidth}
        \includegraphics[width=\linewidth]{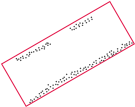}
        \caption{}
    \end{subfigure}
    \caption{H-OBB fitting procedure on 2D projected data. Left: 2D projected input data with convex hull. Centre: Axis-aligned bounding box fitted to each edge rotated parallel to the X-axis. Right: Inverse rotation to minimal bounding box, final result.}
    \label{fig:hobb_fitting}
    \vspace{-1em}
\end{figure}

The H-OBB fitting procedure follows a 2D projection strategy. All segmented points are projected onto the XY-plane, and a convex hull is computed. The convex hull is restricted to a quadrilateral, and each edge is evaluated as a potential candidate for the bounding box length. We rotate each edge to align with the X-axis, compute the minimum-area axis-aligned bounding box, and then select the configuration with the smallest area. The inverse rotation is applied to return the bounding box to its original orientation. This process is visualized in~\Cref{fig:hobb_fitting}.

\subsubsection{Topology Correction} \label{sec:topology_correction}
Even with high-quality plane and bounding box fitting, errors in wall topology often persist, typically caused by 3D scanning noise. We implement a topology-aware correction stage to ensure consistent and realistic wall geometries. Three main correction operations are introduced:

\begin{itemize}
    \item \textbf{Intersection Correction:} Perpendicular bounding boxes that intersect must be clipped or extended to form clean corner connections.
    \item \textbf{Merging:} Collinear bounding boxes with adjacent or overlapping baselines must be merged into longer walls to reduce redundancy.
    \item \textbf{Redundancy Removal:} Smaller boxes fully enclosed by larger ones are treated as artifacts and removed.
\end{itemize}

\begin{figure}[htbp]
    \centering
    \begin{subfigure}[c]{0.35\linewidth}
        \includegraphics[width=\linewidth]{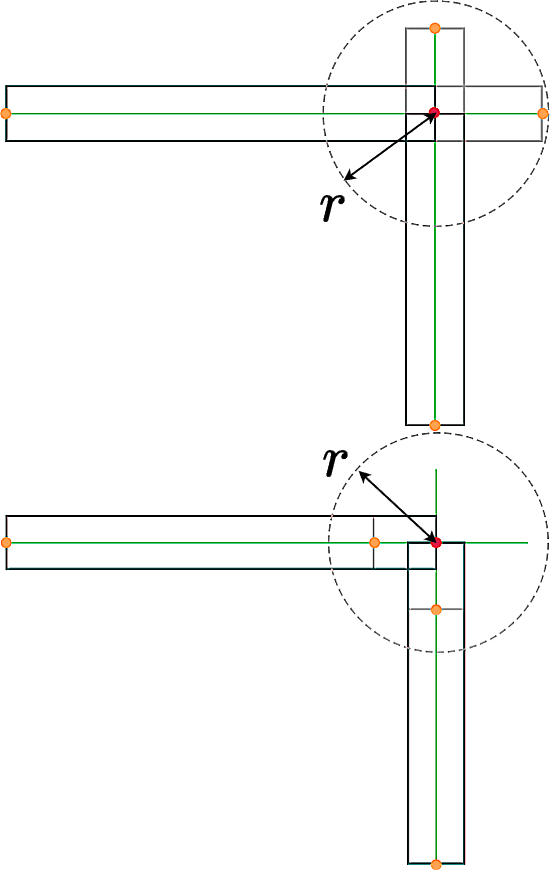}
        \caption{}
        \label{fig:intersection_correction}
    \end{subfigure}
    \hspace{0.5em}
    \begin{subfigure}[c]{0.35\linewidth}
        \includegraphics[width=\linewidth]{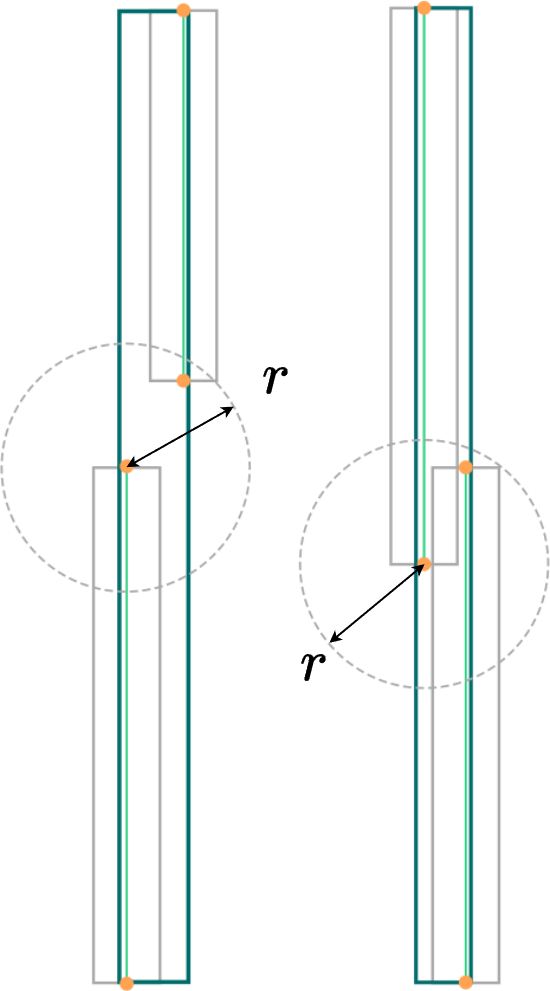}
        \caption{}
        \label{fig:bbox_merging}
    \end{subfigure}
    \caption{Wall bounding box refinement. (a) Extending and clipping bounding boxes. (b) Merging bounding boxes.}
    \vspace{-1em}
\end{figure}

Corrections are based on geometric analysis (adjacency, parallelism, orthogonality) as well as domain knowledge. Perpendicular pairs are found by analysing centrelines of all bounding boxes and computing intersection points. If the endpoint of a bounding box lies within a certain distance $r$ of an intersection, the box is clipped or extended accordingly as shown in~\Cref{fig:intersection_correction}.

For merging, bounding boxes with endpoints within a distance threshold are joined by averaging the height and Z-coordinates of the endpoints while preserving horizontal alignment as shown in~\Cref{fig:bbox_merging}. Intersection correction and merging procedures are applied iteratively, as each operation may introduce new merge or correction candidates. The distance thresholds for these operations (e.g., intersection radius~$r$, merge distance) are set once per project based on domain knowledge such as typical wall widths and corridor dimensions.

\begin{figure}[htbp]
    \centering
    \begin{subfigure}[c]{0.3\linewidth}
        \includegraphics[width=\linewidth]{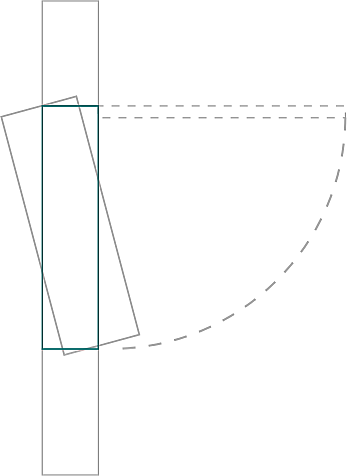}
        \caption{}
        \label{fig:door_projection}
    \end{subfigure}
    \hspace{0.5em}
    \begin{subfigure}[c]{0.5\linewidth}
        \includegraphics[width=\linewidth]{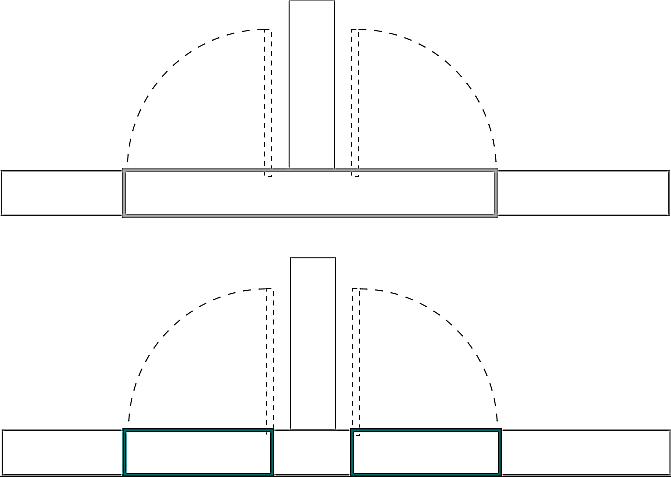}
        \caption{}
        \label{fig:door_splitting}
    \end{subfigure}
    \caption{Bounding box refinement. (a) projecting the bounding box into the parent geometry. (b) splitting bounding boxes too wide.}
    \label{fig:bbox_door_correction}
    \vspace{-1em}
\end{figure}

Finally, to refine door geometry, we project door bounding boxes into their parent wall to ensure alignment as seen in~\Cref{fig:door_projection}. For oversized doors, which may arise from clustering errors, we apply a width threshold and split them into multiple instances. \Cref{fig:door_splitting} shows the corrected doors after offsets were applied to ensure a realistic spacing between the door instances.

These algorithms yield a geometrically and topologically coherent representation of walls and doors, forming a robust foundation for accurate IFC models.

\section{Evaluation} \label{sec:Evaluation}

The primary method for assessing reconstruction accuracy is to compare reconstructed BIMs against ground truth models. While simple cuboid representations can be directly compared, walls and doors may be reconstructed as more complex geometries, making standard geometric comparison difficult. Therefore, we use several evaluation metrics to quantify reconstruction accuracy.

\subsection{Metrics} \label{sec:Metrics}

The main metric used is the 3D-Intersection over Union (3D-IoU). However, due to its limitations in handling small spatial misalignments, we introduce the Volumetric Intersection over Union (vIoU) as an alternative. 

\subsubsection{3D-Intersection over Union}

The 3D-Intersection over Union (3D-IoU) is calculated as the ratio of the volume of the intersection to the volume of the union of the reconstructed and ground-truth bounding boxes. This metric captures both shape and spatial alignment, providing a compact representation of how well two volumes overlap. However, 3D-IoU is highly sensitive to minor deviations in size or alignment: for thin elements such as walls, an offset on the order of the smallest dimension (e.g., wall thickness) already drives the intersection-to-union ratio well below $0.5$, even when the reconstruction is otherwise correct. Furthermore, 3D-IoU requires instance-level assignment between predicted and ground-truth elements, which becomes ambiguous when walls are fragmented or merged during reconstruction.

\subsubsection{Volumetric Intersection over Union}

\begin{figure}[htbp]
    \centering
    \includegraphics[width=0.95\linewidth]{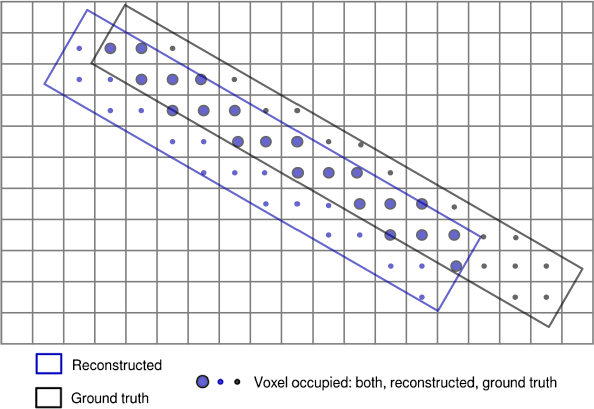}
    \caption{Volumetric IoU computation. Each voxel is classified as belonging to the reconstructed or ground-truth bounding box if its centroid lies within the corresponding geometry.}
    \label{fig:volumetric_iou}
    \vspace{-1em}
\end{figure}

To address these limitations, we propose Volumetric Intersection over Union (vIoU), computed on a voxel grid. Voxel-based IoU is well established in 3D object reconstruction and occupancy prediction~\cite{agnew2021amodal}, where it evaluates per-object shape accuracy. However, its direct application to Scan-to-BIM is non-trivial, as standard per-object voxel IoU inherits the same matching dependency as 3D-IoU. Our vIoU addresses this by operating at the class level: all reconstructed geometry of a given type (e.g., walls) and all corresponding ground-truth geometry are voxelized jointly, and overlap is computed holistically without instance assignment. This is conceptually analogous to how semantic segmentation IoU evaluates per-class predictions in 2D, but applied to 3D volumetric BIM evaluation. We use a voxel size of $5~cm$, corresponding to a tolerance of $\pm2.5~cm$, which we consider reasonable for the structural elements addressed here. The 3D space is discretized, and a voxel is marked as occupied if its centroid lies inside a bounding box. For both ground truth and reconstructed geometries, we count all occupied voxels and compute vIoU as the ratio of their intersection and union. An example is illustrated in~\Cref{fig:volumetric_iou}.

This voxel-based approach simplifies comparison by avoiding complex geometric overlap calculations, provides a sub-voxel tolerance to small misalignments, particularly relevant for thin elements such as walls, and eliminates the need for instance-level bounding box matching. The latter is particularly important for walls, which may appear as a single bounding box, be split into segments, or be fragmented due to incomplete scans. Unlike 3D-IoU, which would assign ground truth to only one predicted instance, vIoU evaluates overlap holistically at the voxel level.

\subsection{Results}

We evaluate our reconstruction pipeline on two datasets: Our newly proposed German Hospital dataset (DeKH) and the CV4AEC Scan-to-BIM Challenge dataset.
While DeKH features largely empty interiors, CV4AEC is fully furnished featuring a lot of clutter and complex occlusions. Furthermore, an ablation study is presented to investigate the impact of individual components in our pipeline and assess the effect of semantic segmentation accuracy.

\subsubsection{CV4AEC} \label{sec:cv4aec}

\begin{table*}[htbp]
    \caption{Results of the evaluation on the CV4AEC dataset
    test scenes. The vIoU is calculated for a 5~cm voxel size.}
    \label{tab:cv4aec_test_scenes}
    \setlength{\tabcolsep}{6pt}
    \small
    \centering
    \begin{tabularx}{\textwidth}{l|cccccc|c|c} \toprule
        \textbf{Metrics} &
        Office\_1 &
        Office\_2 &
        Office\_3 &
        Office\_4 &
        Parking\_1 &
        Parking\_2 &
        \textbf{\textit{Mean \%}} &
        Segmentation IoU \\
        \hline
        Wall 3D-IoU\hfill[-]   & 0.347 & 0.358 & 0.261 & 0.297 & 0.280 & 0.318 & \textit{31.0} & \multirow{2}{*}{0.672} \\ \cline{1-8}
        Wall vIoU\hfill[-]  & 0.437 & 0.383 & 0.259 & 0.302 & 0.386 & 0.420 & \textit{36.5} & ~ \\ \hline
        Door 3D-IoU\hfill[-]   & 0.388 & 0.452 & 0.157 & 0.182 & 0.192 & 0.000 & \textit{22.9} & \multirow{2}{*}{0.586} \\ \cline{1-8}
        Door vIoU\hfill[-]  & 0.367 & 0.407 & 0.150 & 0.149 & 0.247 & 0.000 & \textit{22.0} & ~ \\ \hline
        Column 3D-IoU\hfill[-] & 0.382 & 0.400 & 0.011 & 0.033 & 0.197 & 0.105 & \textit{18.8} & \multirow{2}{*}{0.293} \\ \cline{1-8}
        Column vIoU\hfill[-]     & 0.404 & 0.547 & 0.025 & 0.057 & 0.250 & 0.125 & \textit{23.5} & ~ \\ \hline \hline
        Mean 3D-IoU\hfill[-]   & 0.372 & 0.403 & 0.143 & 0.171 & 0.223 & 0.141 & \textit{24.2} & \multirow{2}{*}{0.517} \\ \cline{1-8}
        Mean vIoU\hfill[-]  & 0.403 & 0.446 & 0.145 & 0.169 & 0.294 & 0.182 & \textit{27.3} & ~ \\ \bottomrule
    \end{tabularx}
    \vspace{-1.2em}
\end{table*}

The CV4AEC dataset
is more challenging due to its cluttered, incomplete scans and presence of interior furniture and occlusions. Six scans from the test set are used for evaluation, with ground-truth BIMs manually created from the input point clouds. Results are presented in~\Cref{tab:cv4aec_test_scenes}.

In the first two scenes, (Office\_1 and Office\_2) with standard wall and door configurations, our pipeline performs best due to alignment with our assumptions. Scenes Office\_3 and Office\_4 contain open spaces, cubicles, and complex facade elements, making the reconstruction more challenging. The last two scenes (Parking\_1 and Parking\_2) are from underground parking structures. Although some improvement in column detection is observed, results remain lower than in the office scenes. Notably, no doors are detected in (Parking\_2), likely due to closed doors during scanning and segmentation failure suggesting that a dedicated detection algorithm for doors might be beneficial. The segmentation IoU presented in the last column gives an indication of the challenge these types of data are posing to state-of-the-art point cloud segmentation methods.

\subsubsection{German Hospital Dataset (DeKH)} \label{sec:dekh}
Our DeKH dataset is a new public dataset including four scans from three buildings in an unused, largely unfurnished hospital building in Germany. Building \textbf{A}, over 100 years old, features structural walls, central corridors, and a symmetrical layout. Building \textbf{B} includes an Intensive Care Unit (ICU) area with integrated equipment, while building \textbf{C} contains empty surgical rooms with built-in furniture.

In addition to the point clouds annotated following a construction ontology-based guideline~\citep{kaufmann2023ontology}, we provide the manually created BIMs for all of the scans. The results of the reconstruction are presented in~\Cref{tab:DeKH_scenes_results}.

\begin{table}[htbp]
    \caption{Results of the evaluation on the DeKH dataset. The vIoU is calculated for a voxel size of 5~cm.}
    \label{tab:DeKH_scenes_results}
    \setlength{\tabcolsep}{3pt} 
    \small
    \centering
    \begin{tabular}{l|cccc|c} \toprule    
        \multirow{2}{*}{\textbf{Metrics}} & \textbf{A} & \textbf{A} & \textbf{B} & \textbf{C} & \textbf{\textit{Mean}} \\
        ~ & 1st floor & 2nd floor & ICU & Surgery & \textbf{\textit{\%}} \\ \hline
        Wall 3D-IoU\hfill[-]  & 0.395 & 0.219 & 0.270 & 0.328 & \textit{30.3} \\ \hline
        Wall vIoU\hfill[-] & 0.553 & 0.463 & 0.388 & 0.451 & \textit{46.4} \\ \hline
        Door 3D-IoU\hfill[-]  & 0.417 & 0.104 & 0.193 & 0.223 & \textit{23.4} \\ \hline
        Door vIoU\hfill[-] & 0.402 & 0.103 & 0.219 & 0.236 & \textit{24.0} \\ \hline \hline
        Mean 3D-IoU\hfill[-]  & 0.406 & 0.162 & 0.232 & 0.275 & \textit{26.9} \\ \hline
        Mean vIoU\hfill[-]      & 0.478 & 0.283 & 0.304 & 0.344 & \textit{35.2} \\ \bottomrule
    \end{tabular}
\end{table}

Performance in the \textbf{A} building's second floor is lower than the first floor. The pipeline faced challenges in reconstructing smaller wall compartments. The door reconstruction had issues due to sparse segmentation. In the \textbf{B} ICU building, performance improves. Door reconstruction is accurate but scored poorly due to a mismatch in door representations between the ground truth of thin door leaves and the reconstructed geometry including frame and lining. In building \textbf{C}, some furniture surfaces were misclassified as walls, and some open doors lead to double reconstructions.

\begin{figure}[htbp]
    \centering
    \hspace{0.5em}
    \begin{subfigure}[c]{\linewidth}
            \begin{subfigure}[c]{0.45\linewidth}
            \includegraphics[width=\linewidth]{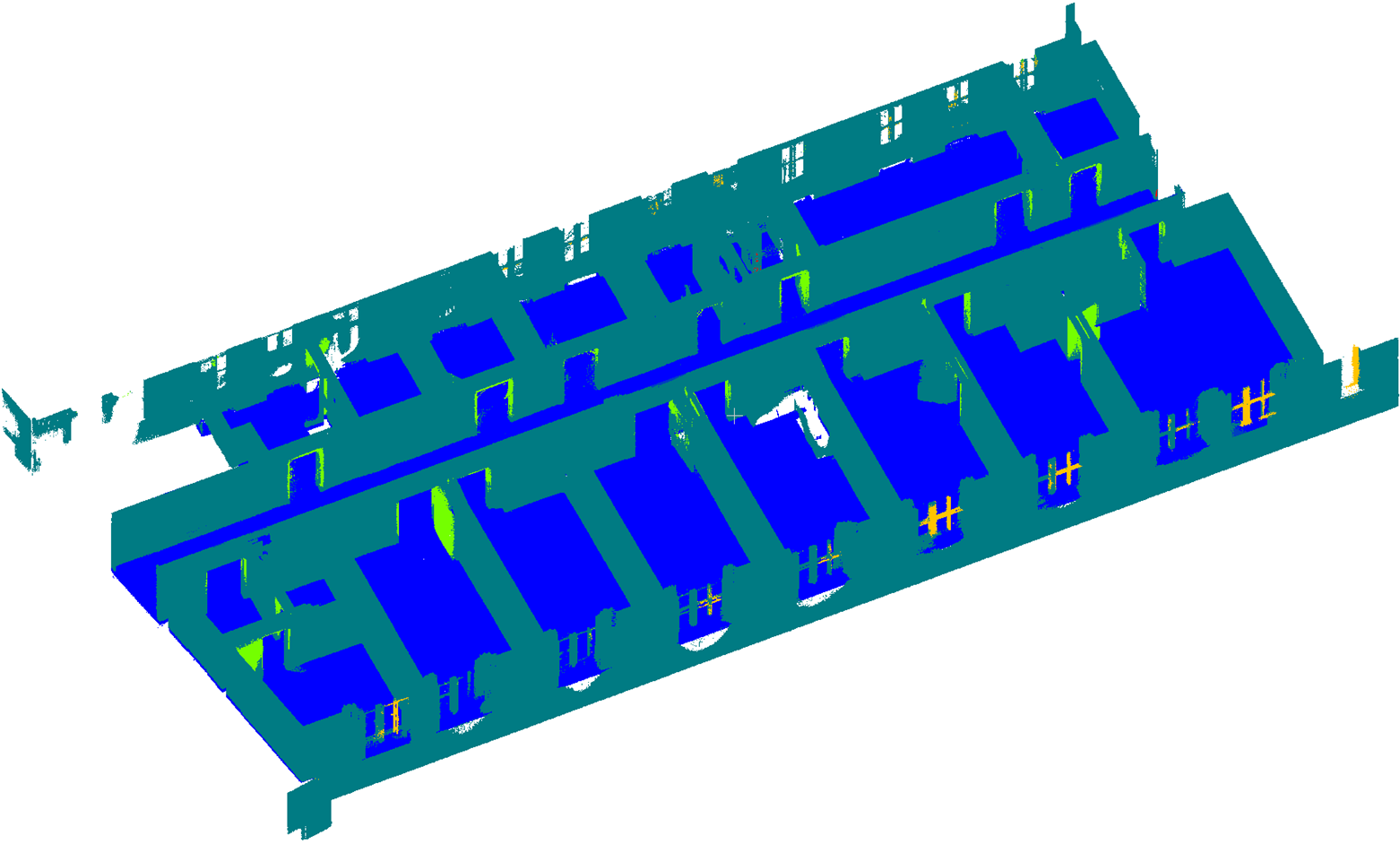}
            \caption*{Semantic segmentation}
        \end{subfigure}
        \hfill
        \begin{subfigure}[c]{0.45\linewidth}
            \includegraphics[width=\linewidth]{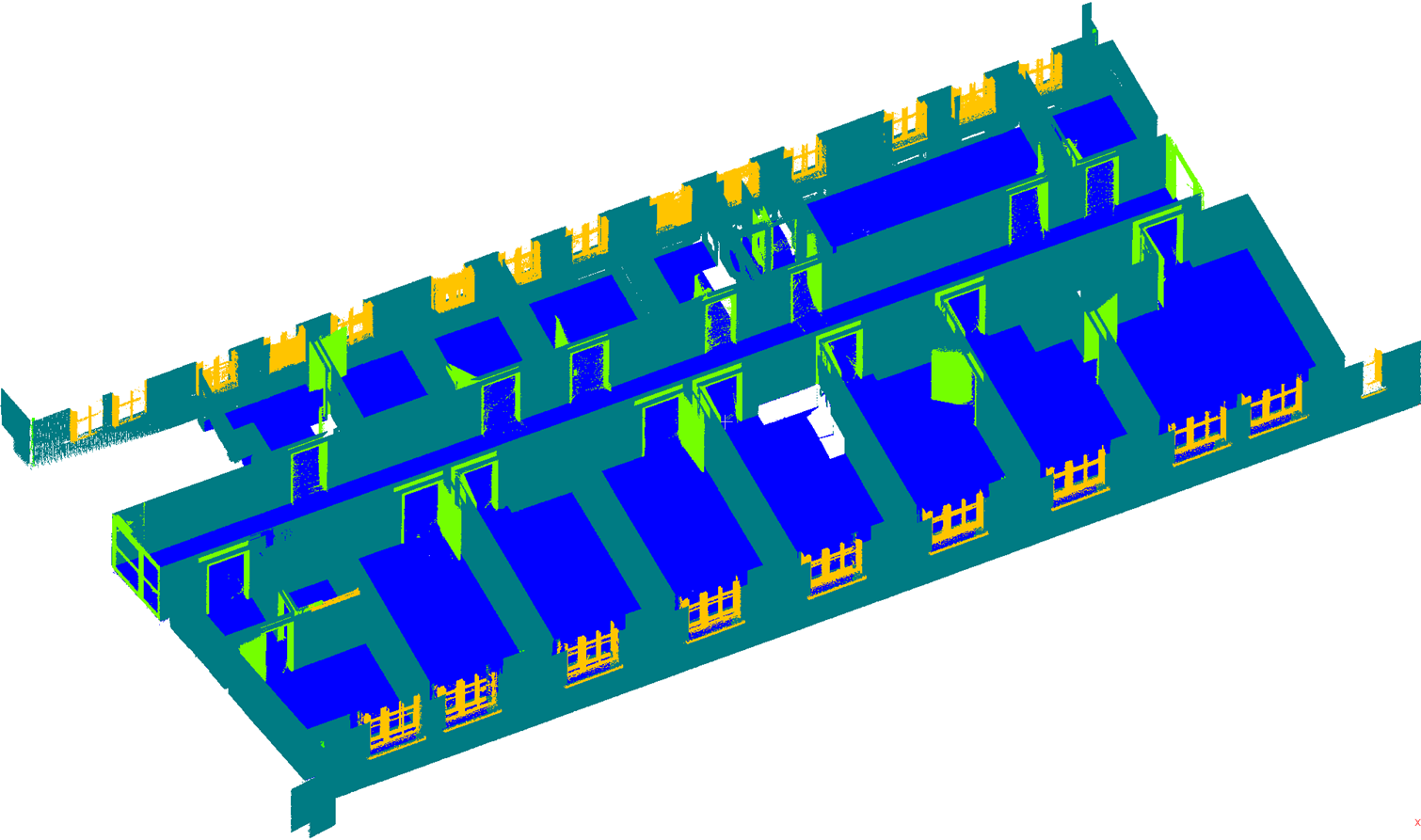}
            \caption*{Ground truth}
        \end{subfigure}
        \begin{subfigure}[c]{0.45\linewidth}
            \includegraphics[width=\linewidth]{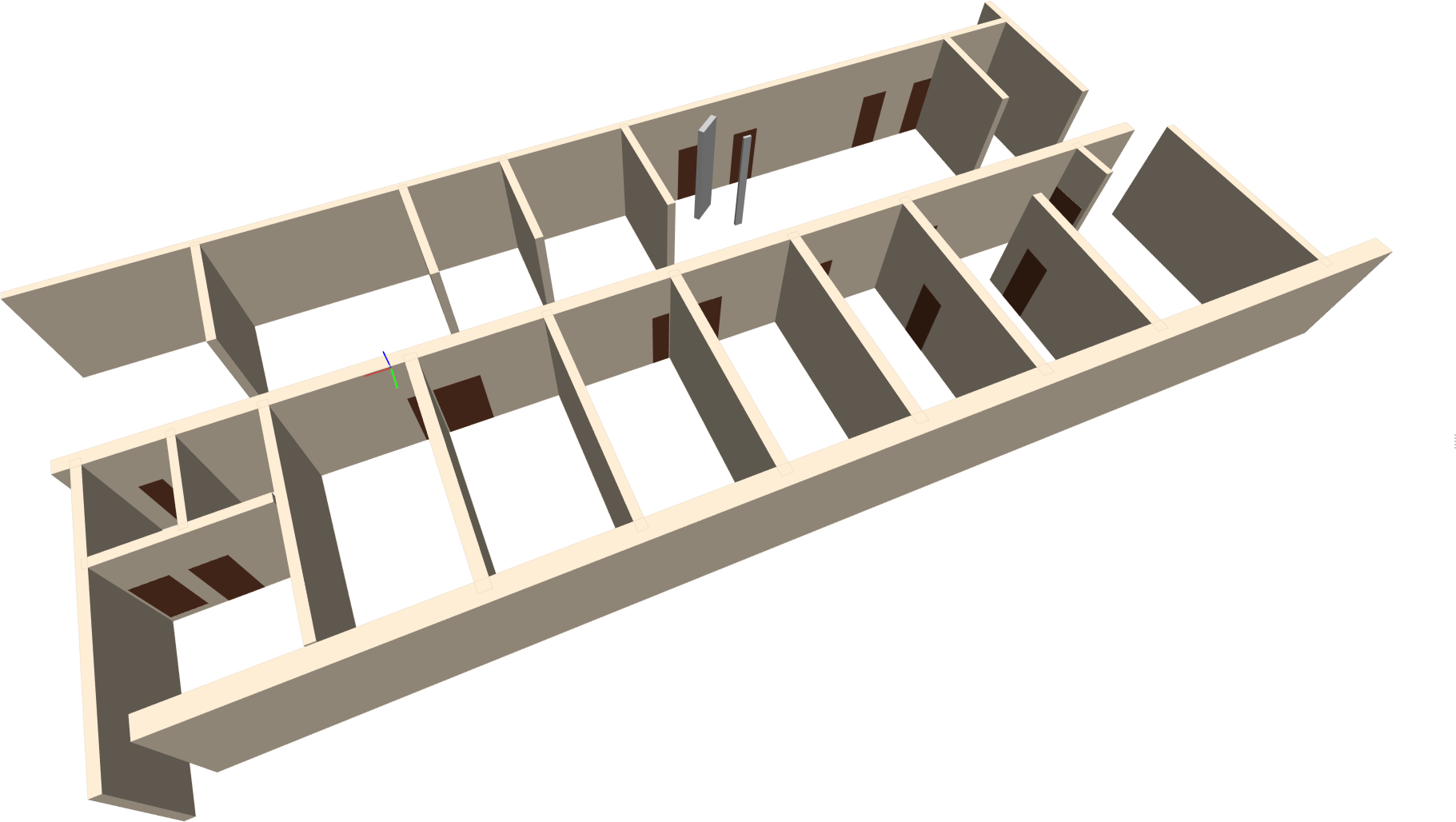}
            \caption*{Reconstructed BIM model from semantic segmentation}
        \end{subfigure}
        \hfill
        \begin{subfigure}[c]{0.45\linewidth}
            \includegraphics[width=\linewidth]{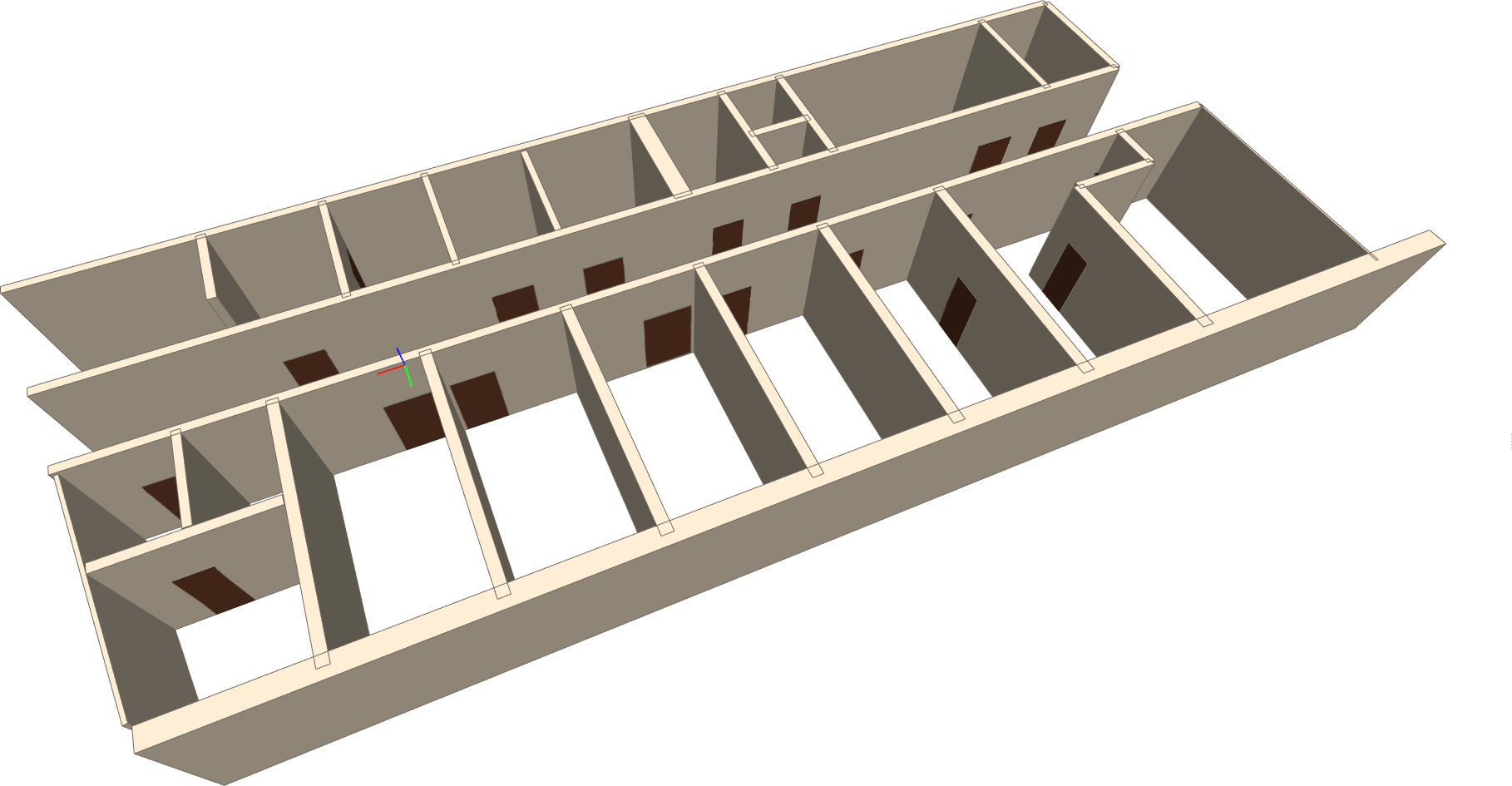}
            \caption*{Reconstructed BIM model from the ground truth labels}
        \end{subfigure}
    \end{subfigure}
    
    \begin{subfigure}[c]{0.45\linewidth}
        \includegraphics[width=\linewidth]{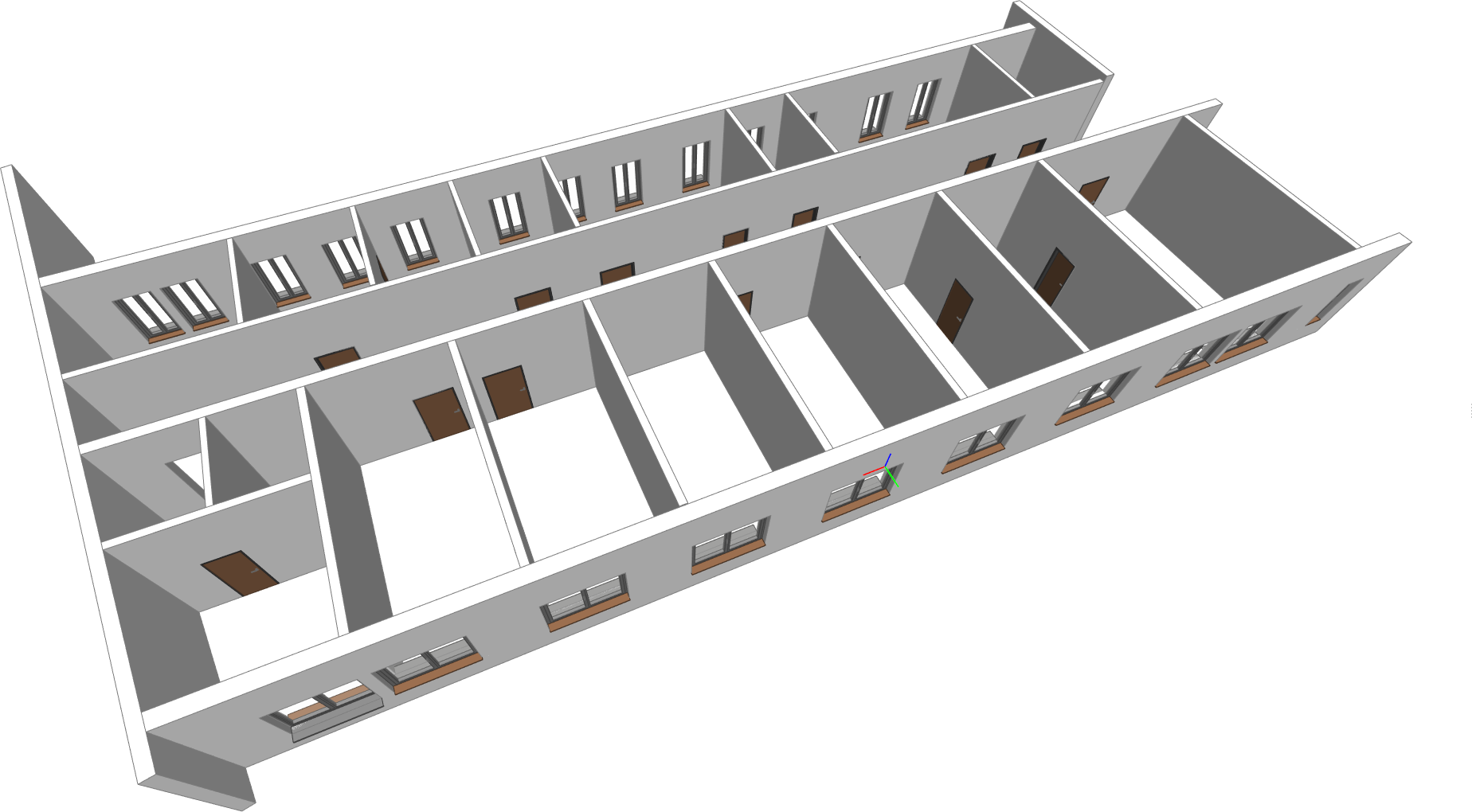}
        \caption*{Ground truth BIM model}
    \end{subfigure}
    
    \caption{Comparison of reconstructed BIM models from semantic segmentation vs ground truth in DeKH--building \textbf{A}.}
    \label{fig:DeKH_A_qualitative}
    \vspace{-0.5em}
\end{figure}

The ablation study in~\Cref{tab:DeKH_A_1st_ablation} presents three configurations: \textbf{Baseline}, which follows the reconstruction approach of~\cite{kaufmann2022scalebim} using RANSAC for primitive fitting and H-OBB estimation without topology correction; \textbf{Ours}, representing the full pipeline with HYSAC, H-OBB fitting, and topology refinement; and \textbf{GT labels}, which applies the same method as \textbf{Ours} but uses ground truth semantic labels to isolate the effect of segmentation quality.

The results confirm our pipeline significantly improves wall and door reconstruction accuracy compared to the baseline. Notably, even with ground truth labels, wall reconstruction does not improve. Our approach achieves a $4.9\%$ higher bounding box IoU while volumetric IoU for walls shows only a slight advantage with ground truth labels. The door reconstruction improves considerably with ground truth segmentation. These results highlight that, while our method can tolerate some incomplete segmentation especially for large structures like walls, accurate semantic segmentation remains crucial for smaller elements like doors. The qualitative results in~\Cref{fig:DeKH_A_qualitative} also support these findings.

\begin{table}[htbp]
    \caption{Ablation study on the first floor of the DeKH \textbf{A} building.}
    \label{tab:DeKH_A_1st_ablation}
    \setlength{\tabcolsep}{8pt}
    \small
    \centering
    \begin{tabular}{l|cc|c} \toprule
        \textbf{Metrics}     & \textbf{Baseline} & \textbf{Ours}  & \textbf{GT Labels} \\ \hline
        Wall 3D-IoU\hfill[-] & 0.371             & \textbf{0.395} & 0.346 \\ \hline
        Wall vIoU\hfill[-]   & 0.466             & \textbf{0.553} & 0.571 \\ \hline
        Door 3D-IoU\hfill[-] & 0.322             & \textbf{0.417} & 0.519 \\ \hline
        Door vIoU\hfill[-]   & 0.391             & \textbf{0.402} & 0.521 \\ \hline \hline
        Mean 3D-IoU\hfill[-] & 0.346             & \textbf{0.406} & 0.432 \\ \hline
        Mean vIoU\hfill[-]   & 0.429             & \textbf{0.478} & 0.546 \\ \bottomrule
    \end{tabular}
    \vspace{-1em}
\end{table}

\section{Discussion and Conclusion} \label{sec:Conclusion}

This work presents a robust and modular pipeline that advances the state of scan-to-BIM automation, generating IFC-compliant BIM models from 3D point clouds. The proposed pipeline achieves competitive reconstruction accuracy across diverse scenes, demonstrating its practical viability and scalability. We contribute to the topic with methodological advances and a new benchmark dataset.

Our approach delivers accurate reconstructions of structural elements using a hybrid methodology of learning-based and geometric processing techniques. Evaluation across two datasets confirms the pipeline's robustness and effectiveness, particularly for walls and doors, where reconstruction quality exceeds a RANSAC-based baseline. Even with incomplete or noisy data, the system maintained reliable performance. Direct comparison with existing scan-to-BIM methods is currently not feasible due to the absence of open-source implementations and standardized evaluation protocols. The CV4AEC challenge provides the closest available shared benchmark, where the proposed method was evaluated competitively. The release of DeKH aims to further support reproducible comparison in future work.

A major contribution is the introduction of the \textbf{DeKH} benchmark dataset. The release of DeKH, with high-resolution scans, ground truth BIMs, and semantic annotations, provides a new benchmark for scan-to-BIM.

Additionally, we propose vIoU, which adapts voxel-based IoU to class-level evaluation without instance matching, providing more robust comparison in scenes with misaligned, fragmented, or variably segmented elements.

To strengthen the pipeline, future efforts should expand beyond the current object classes toward more functional BIMs. The reconstruction algorithms are adaptable and could extend to elements such as windows, stairs, and mechanical, electrical, and plumbing (MEP) systems.

The topology refinement procedures showed sensitivity to parameter selection. For instance, these operations can introduce inaccuracies when small segments are wrongly merged. Thus, adaptive, context-aware refinement strategies or learned alternatives could address this vulnerability.

Our pipeline adopts the Manhattan world assumption, which constrains reconstructed geometry to predominantly orthogonal orientations. Consequently, it is best suited to buildings with rectilinear layouts, such as offices, hospitals, and residential complexes, which constitute the majority of typical building stock. Structures with significant non-orthogonal elements (such as curved walls, angled fa\c{c}ades, or irregular floor plans) would not be handled correctly, as the axis-aligned clustering would either discard or mis-assign their points, and the bounding box fitting would produce poor approximations. Relaxing this assumption, for instance through data-driven orientation estimation or by supporting polygonal wall primitives, is an important direction for future work.

Looking ahead, extending the pipeline to support a broader range of object classes and incorporating subcomponent-level reasoning will be important steps toward producing complete and functionally rich BIMs. In parallel, research into end-to-end learning approaches that can infer BIM semantics and geometry directly from raw point cloud data and other sources will help move the field forward.

\section*{Acknowledgements}

This research was funded by the European Union as part of the projects: HumanTech (Grant Agreement 101058236) and ShieldBOT (Grant Agreement 101235093).

\section*{References}
\bibliography{references}

\clearpage

\twocolumn[%
  \begin{center}
    {\titlefont Supplementary Material:\\
     BIMStruct3D: A Fully Automated Hybrid Learning Scan-to-BIM Pipeline with Integrated Topology Refinement\par}
    \vspace{8mm}
    {\authorfont Mahdi Chamseddine$^{1,2}$, Fabian Kaufmann$^2$, Marius Schellen$^2$, Christian Glock$^2$,\\
     Didier Stricker$^{1,2}$, and Jason Rambach$^{1}$\par}
    \vspace{4mm}
    {\authorfont $^1$German Research Center for Artificial Intelligence (DFKI), Kaiserslautern, Germany\\
     $^2$RPTU Kaiserslautern-Landau, Kaiserslautern, Germany\par}
  \end{center}
  \vspace{6mm}
]
\thispagestyle{empty}

\setul{2pt}{0.25pt}

\section{CV4AEC Scene Names} \label{sec:appendix_A}

The mapping of the scene names used in~\Cref{tab:cv4aec_test_scenes} is provided in~\Cref{tab:cv4aec_scene_mapping} for reference and potential comparison of results.

\begin{table}[h]
    \caption{CV4AEC dataset~\citep{armeni2024cv4aec} scene name mapping.}
    \label{tab:cv4aec_scene_mapping}
    \centering
    \small
    \begin{tabular}{l|l} \toprule
        \textbf{Short Name} & \textbf{CV4AEC Scene} \\ \hline
        Office\_1 & 08\_ShortOffice\_01\_F1 \\
        Office\_2 & 08\_ShortOffice\_01\_F2 \\
        Office\_3 & 11\_MedOffice\_05\_F2 \\
        Office\_4 & 11\_MedOffice\_05\_F4 \\
        Parking\_1 & 25\_Parking\_01\_F1 \\
        Parking\_2 & 25\_Parking\_01\_F2 \\ \bottomrule
    \end{tabular}
\end{table}

The CV4AEC challenge comprises of two challenges, a 2D and 3D one. Here we use the test data of the 3D challenge for evaluation. The training data of the challenge was used for training the point cloud segmentation model.

\section{DeKH Dataset} \label{sec:appendix_B}

\begin{table*}[bp]
    \caption{Overview of the DeKH dataset, including colored point clouds, semantic labels, and ground truth BIM models}
    \centering
    \renewcommand{\arraystretch}{4} 
    \setlength{\tabcolsep}{6pt}    
    \vspace{-2em}

    \begin{tabular}{
        >{\centering\arraybackslash}m{0.5cm}
        >{\centering\arraybackslash}m{3.4cm} 
        >{\centering\arraybackslash}m{3.4cm} 
        >{\centering\arraybackslash}m{3.4cm} 
        >{\centering\arraybackslash}m{3.4cm}
    }
        & \textbf{Building A - 1st floor} & \textbf{Building A - 2nd floor} & \textbf{Building B - ICU} & \textbf{Building C - Surgery} \\
        \hline
        
        \rotatebox{90}{\textbf{PCD RGB}} &
        \includegraphics[width=3.4cm]{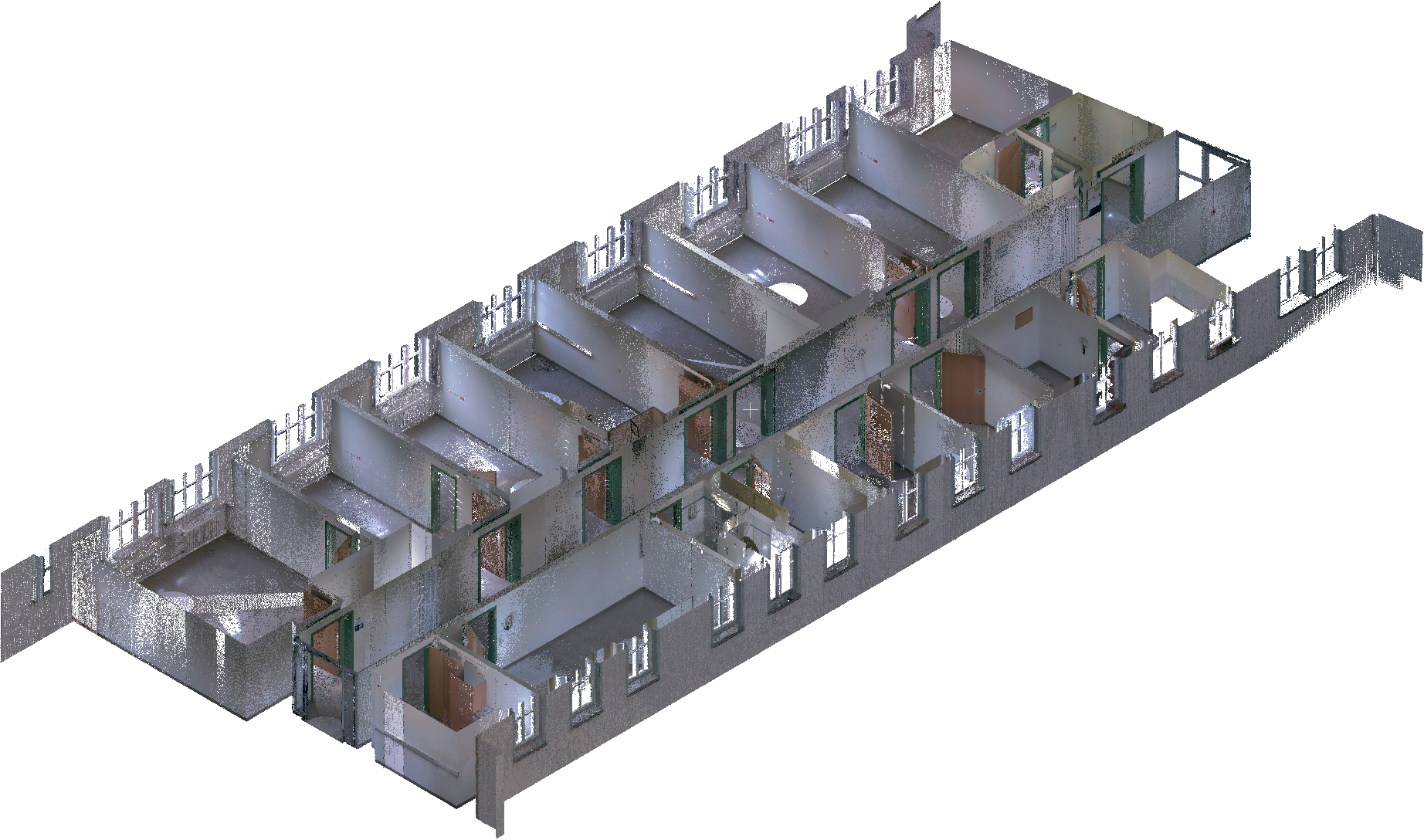} &
        \includegraphics[width=3.4cm]{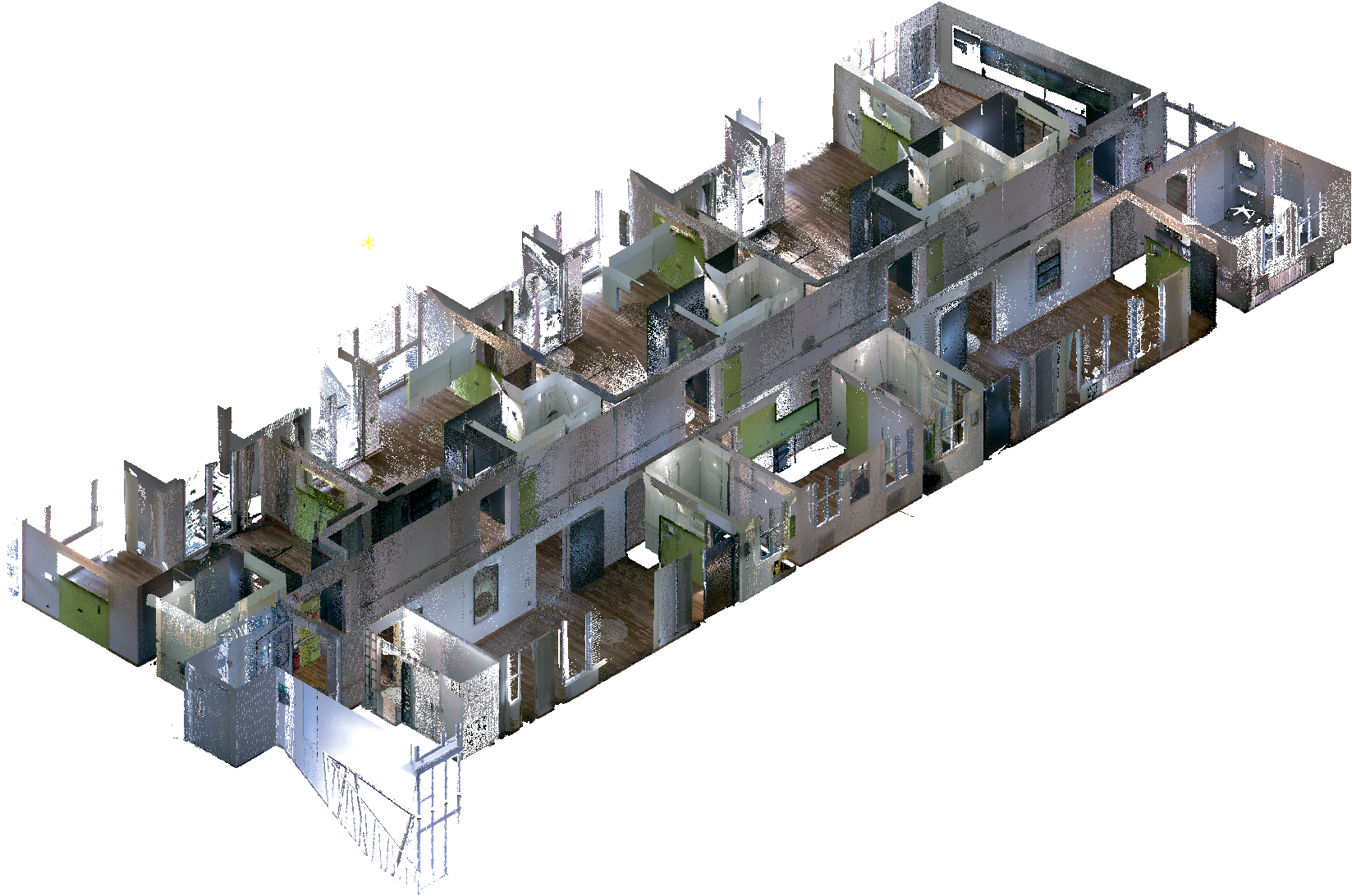} &
        \includegraphics[width=3.4cm]{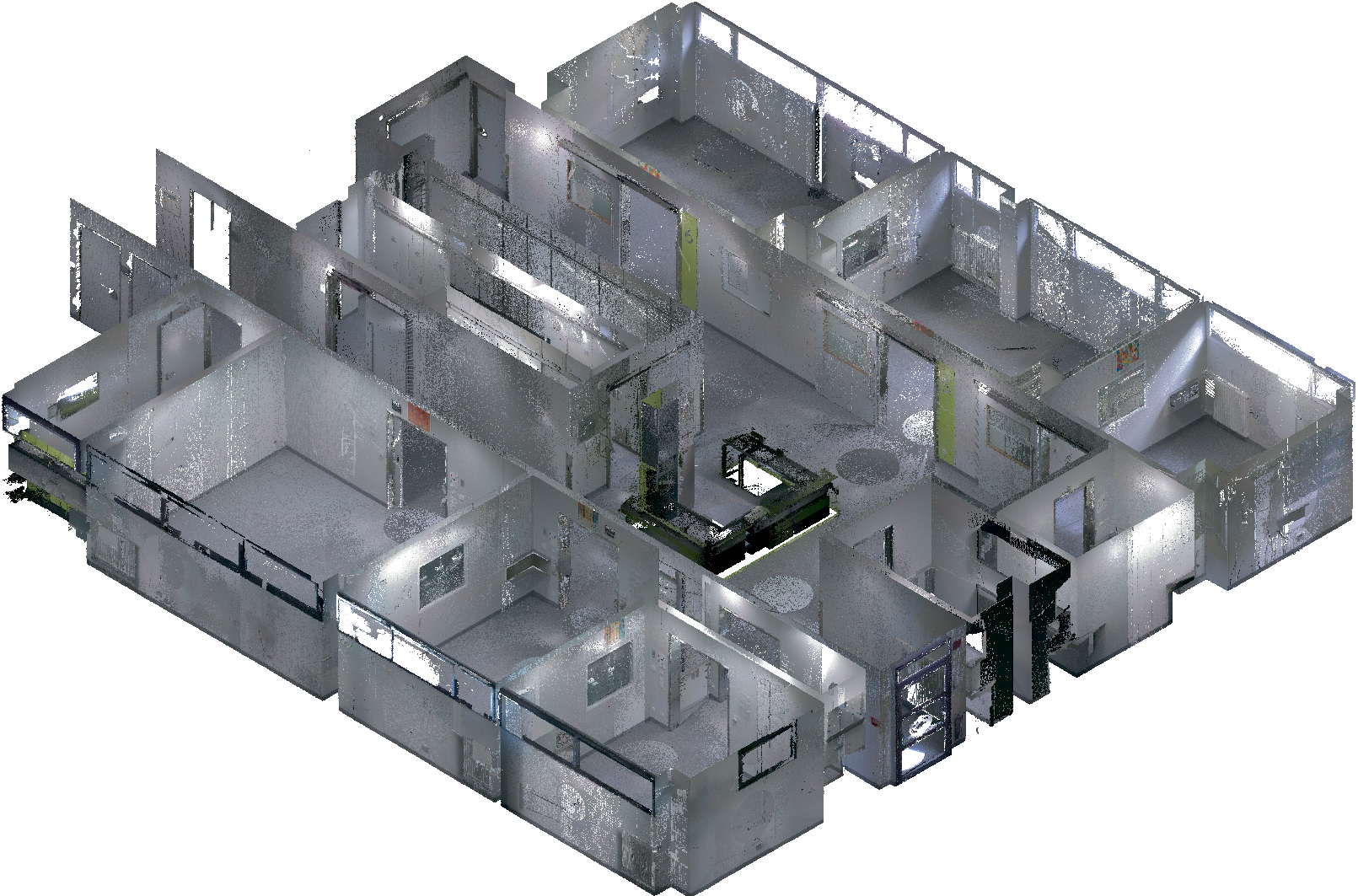} &
        \includegraphics[width=3.4cm]{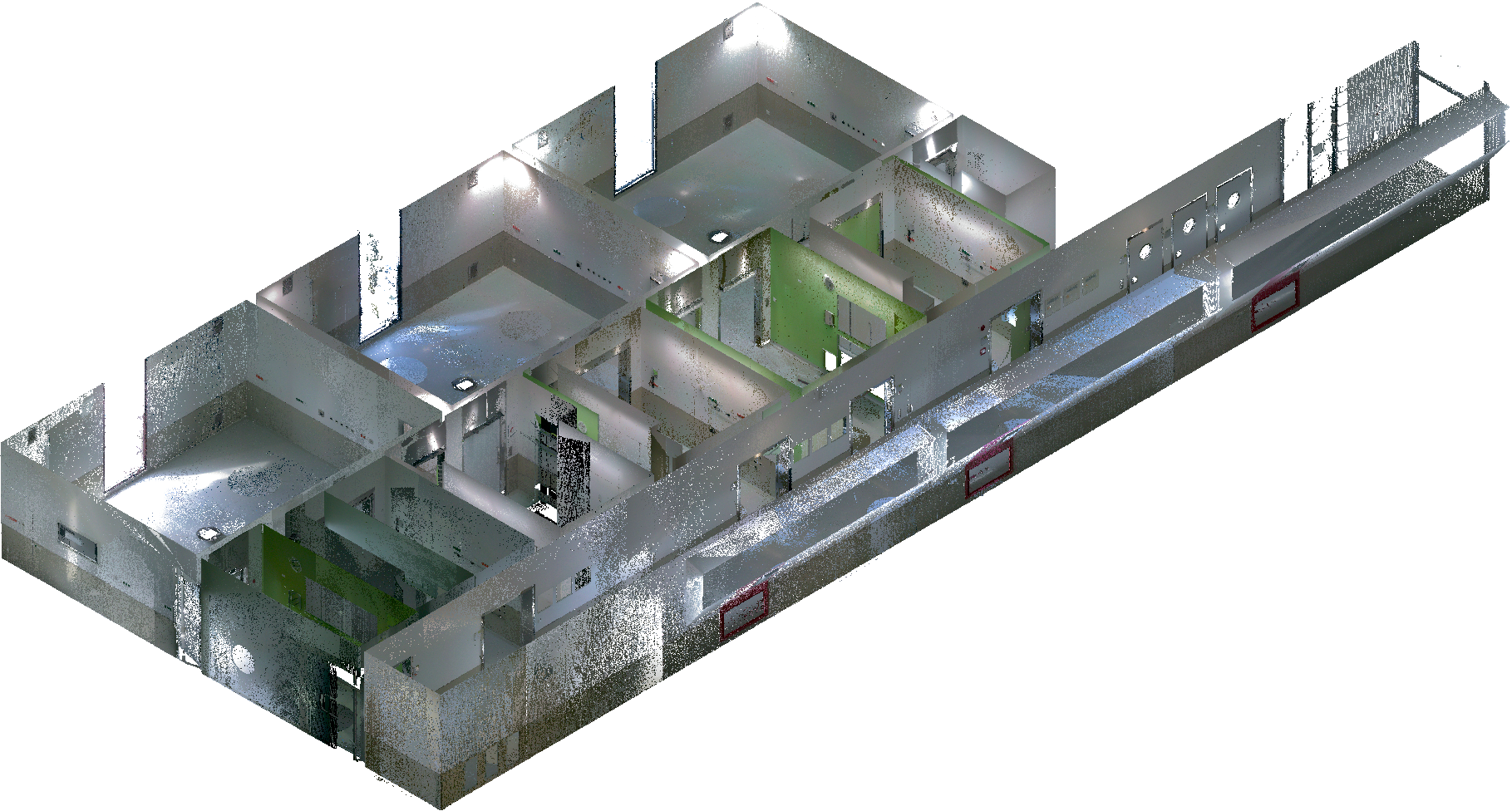} \\
        
        \rotatebox{90}{\textbf{PCD labeled}} &
        \includegraphics[width=3.4cm]{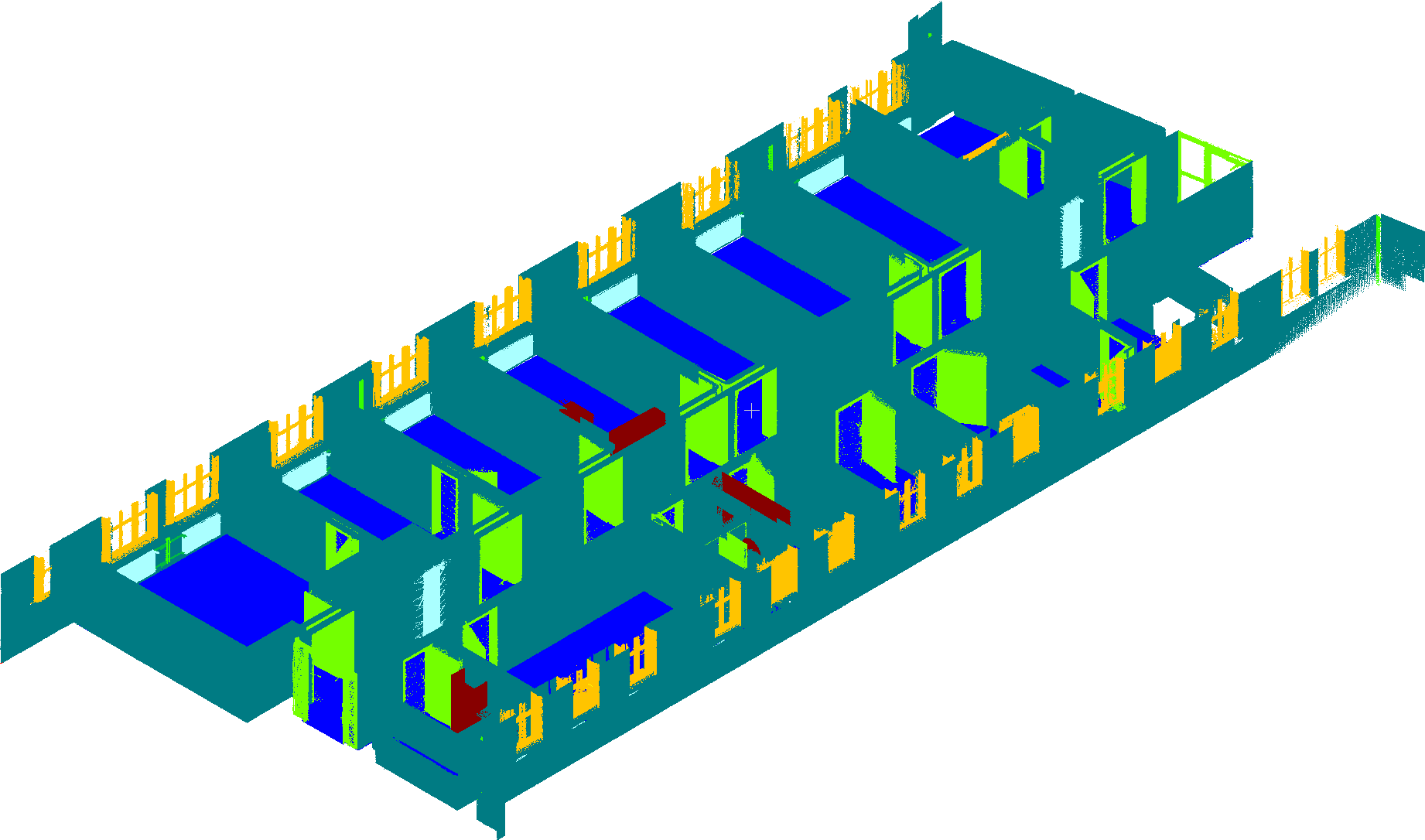} &
        \includegraphics[width=3.4cm]{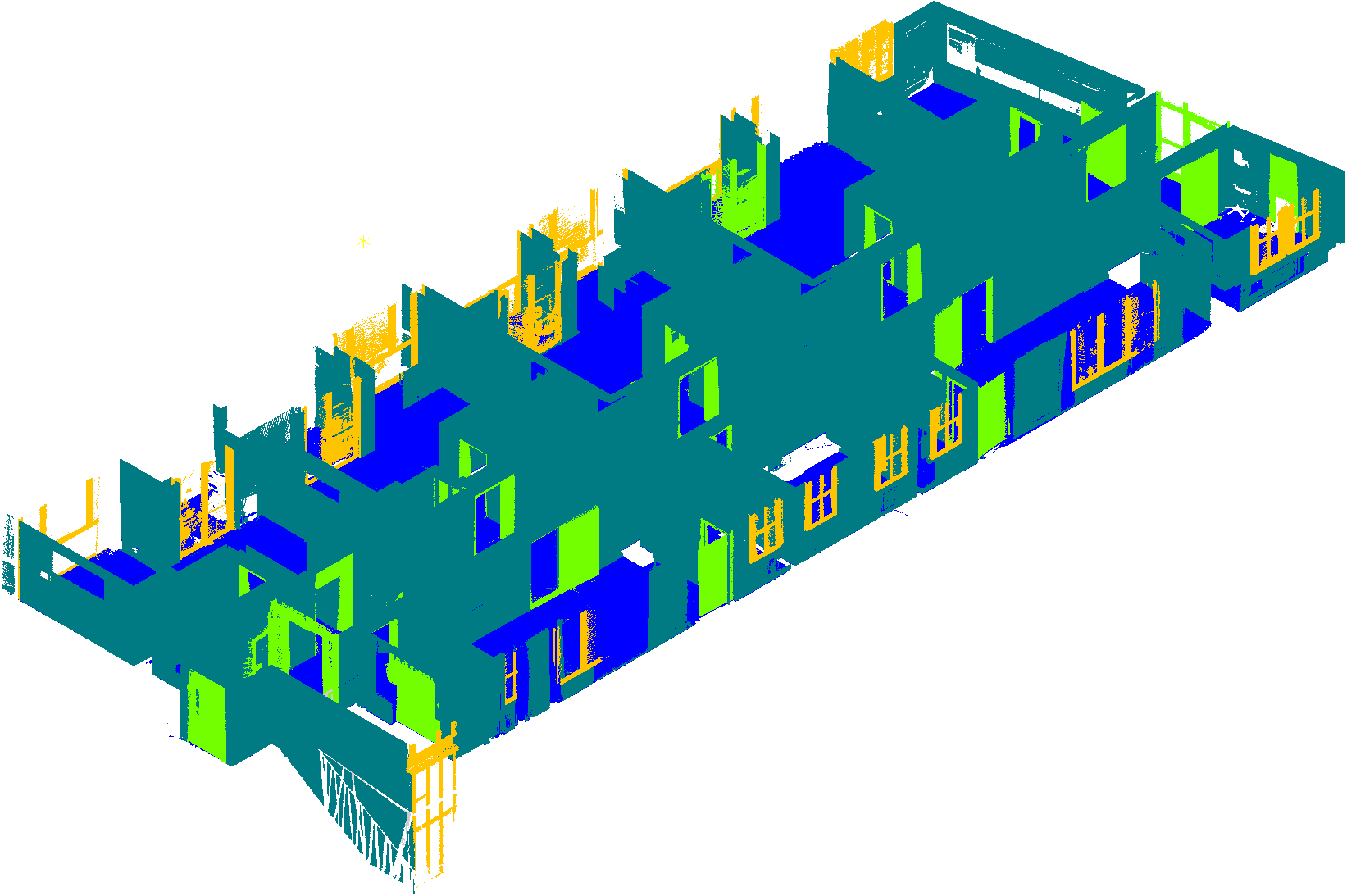} &
        \includegraphics[width=3.4cm]{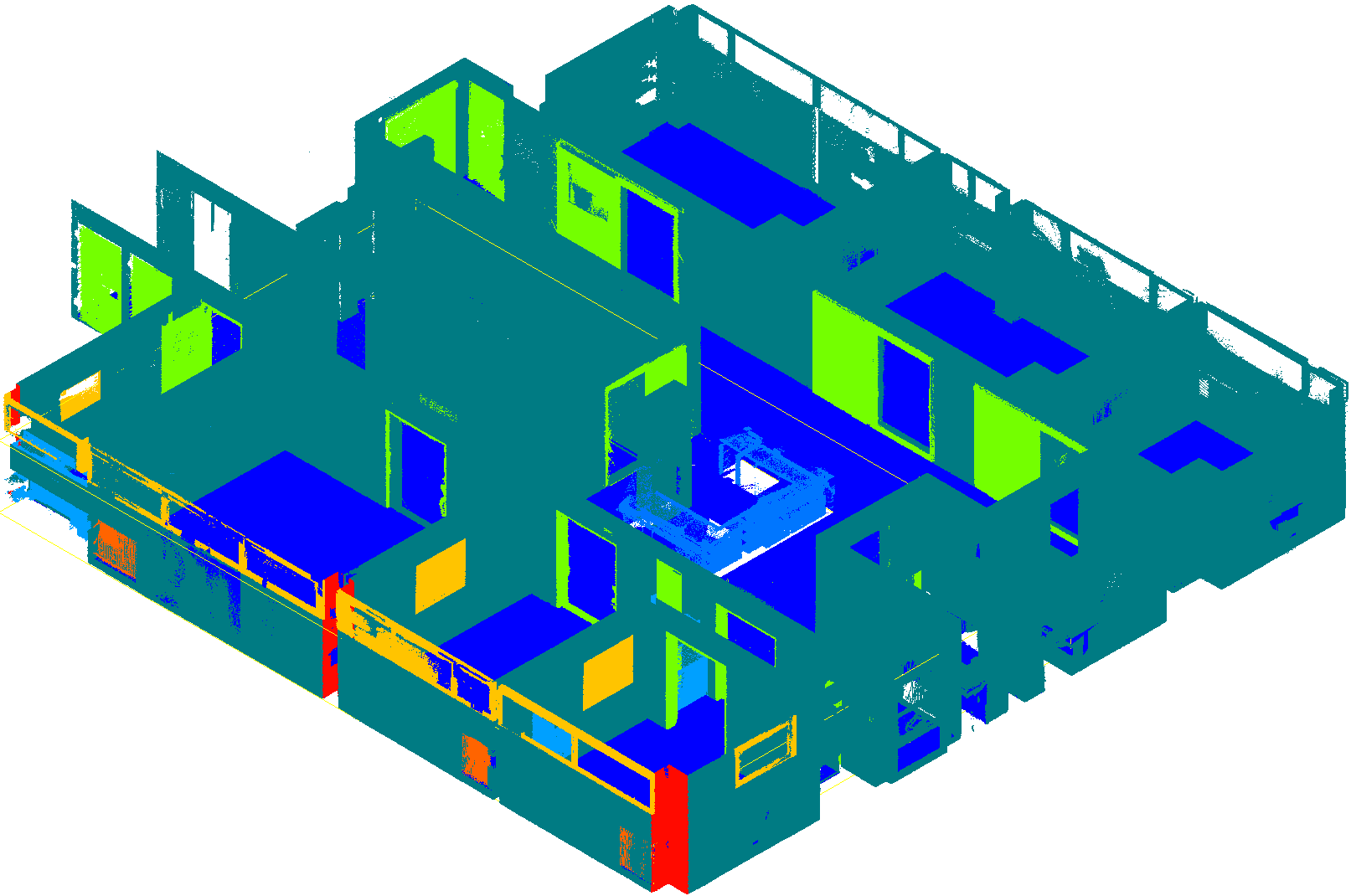} &
        \includegraphics[width=3.4cm]{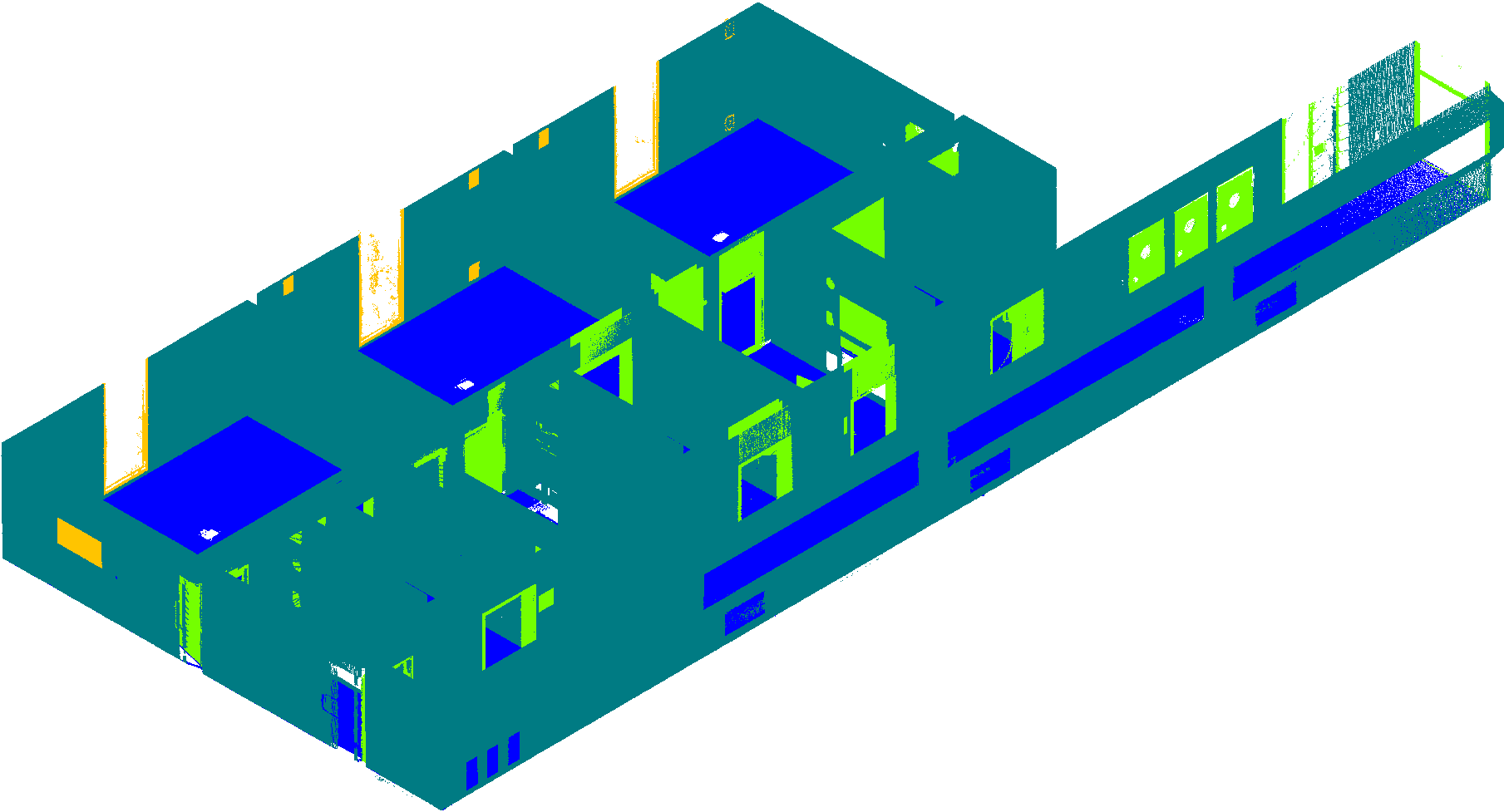} \\
        
        \rotatebox{90}{\textbf{GT BIM}} &
        \includegraphics[width=3.4cm]{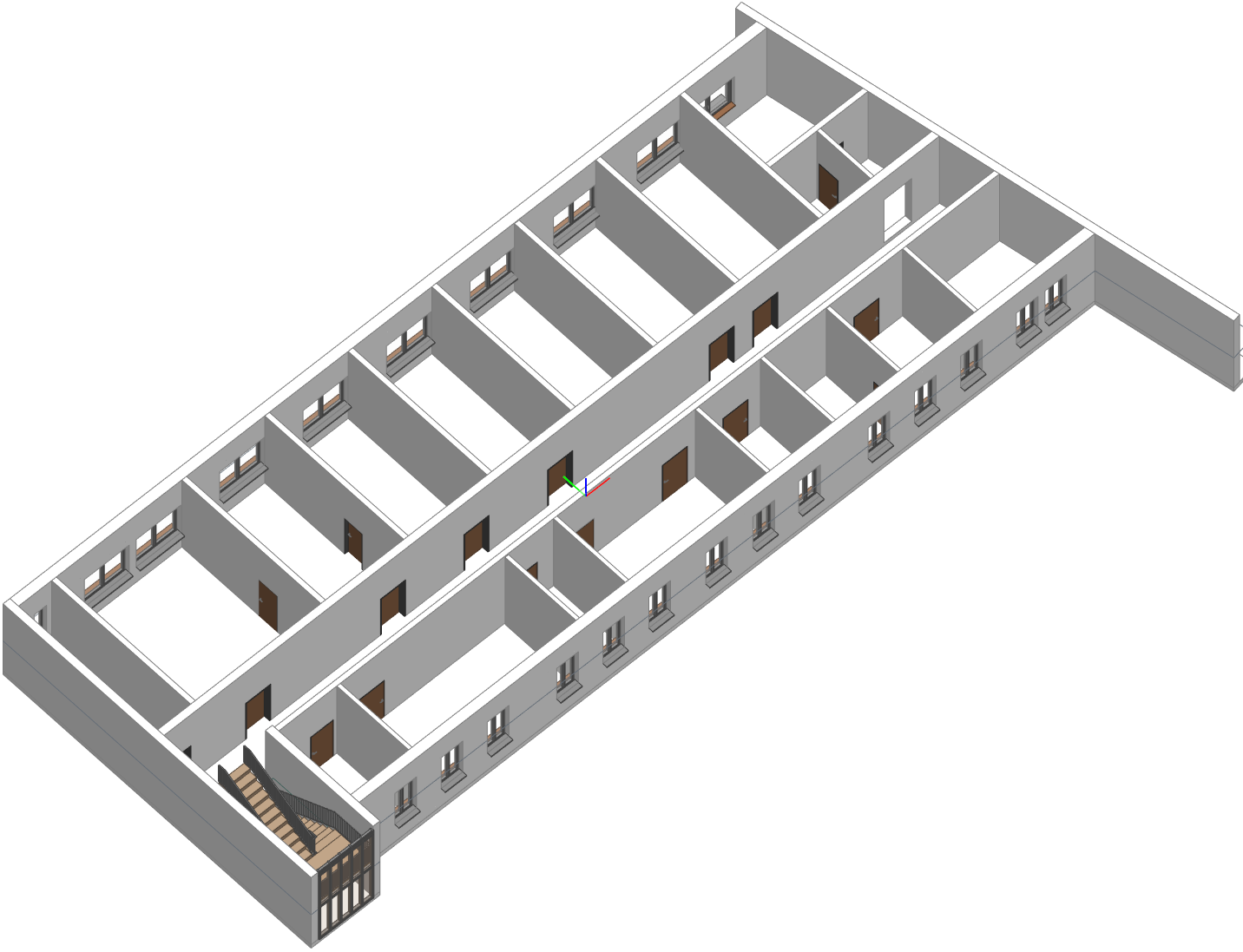} &
        \includegraphics[width=3.4cm]{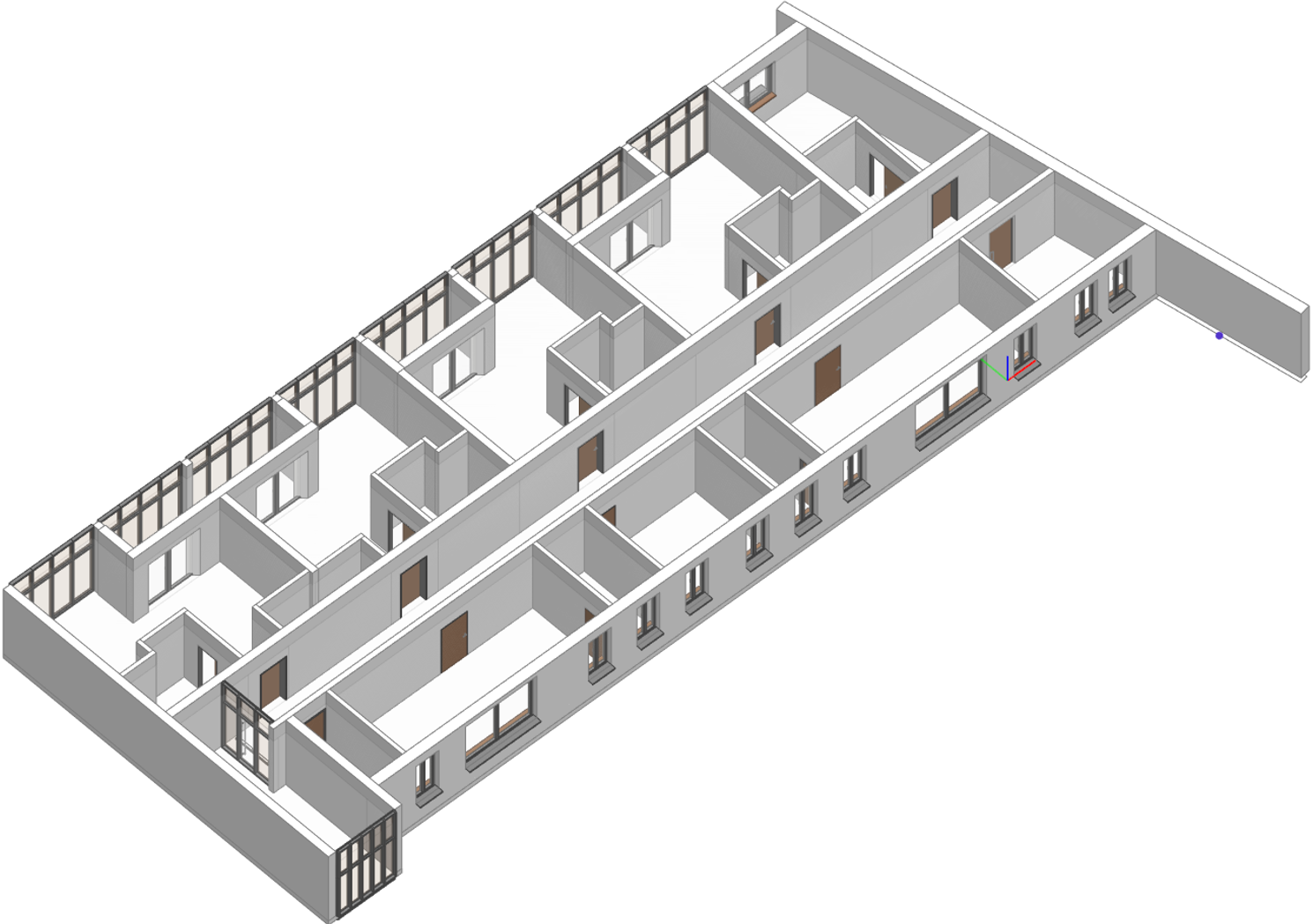} &
        \includegraphics[width=3.4cm]{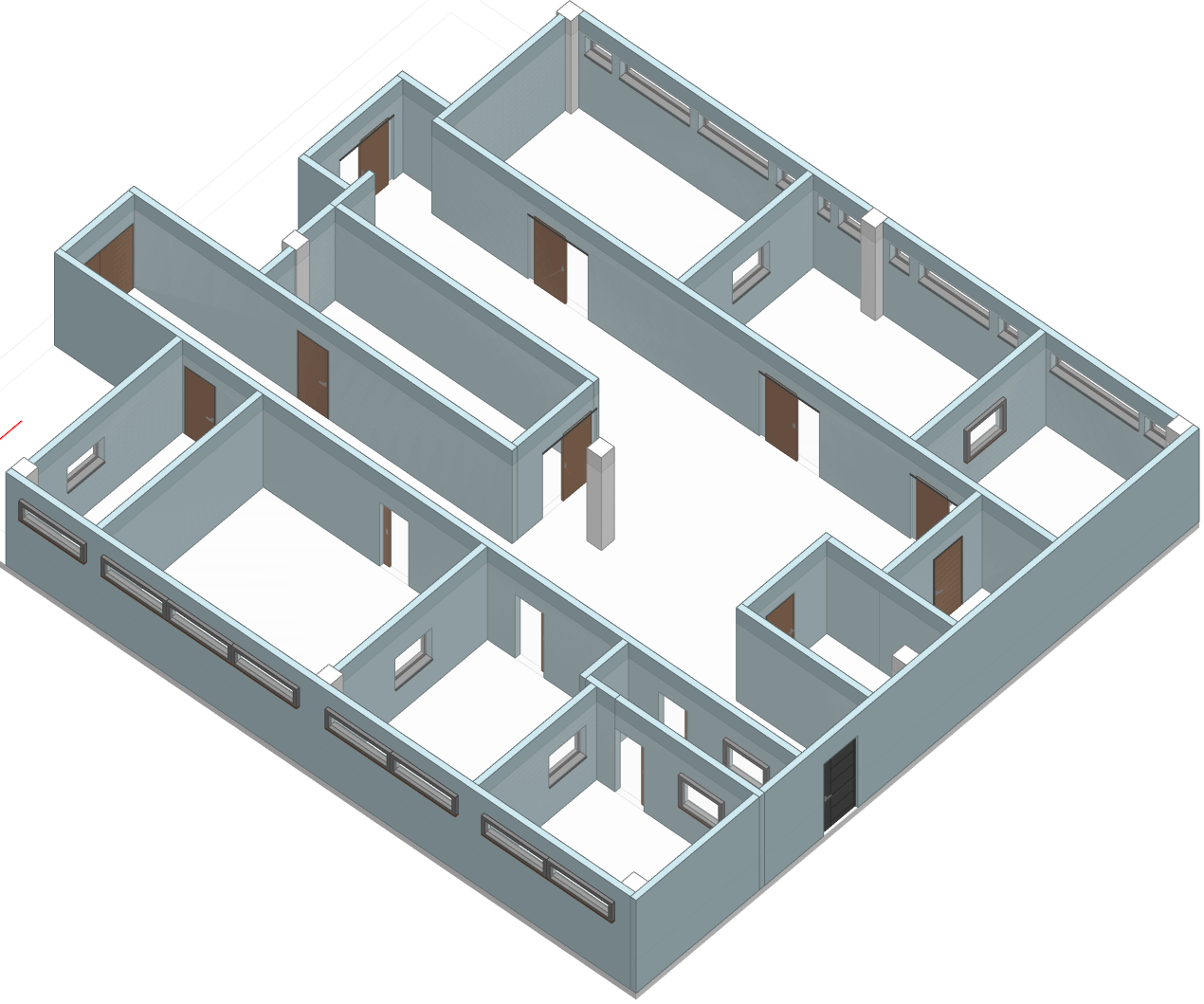} &
        \includegraphics[width=3.4cm]{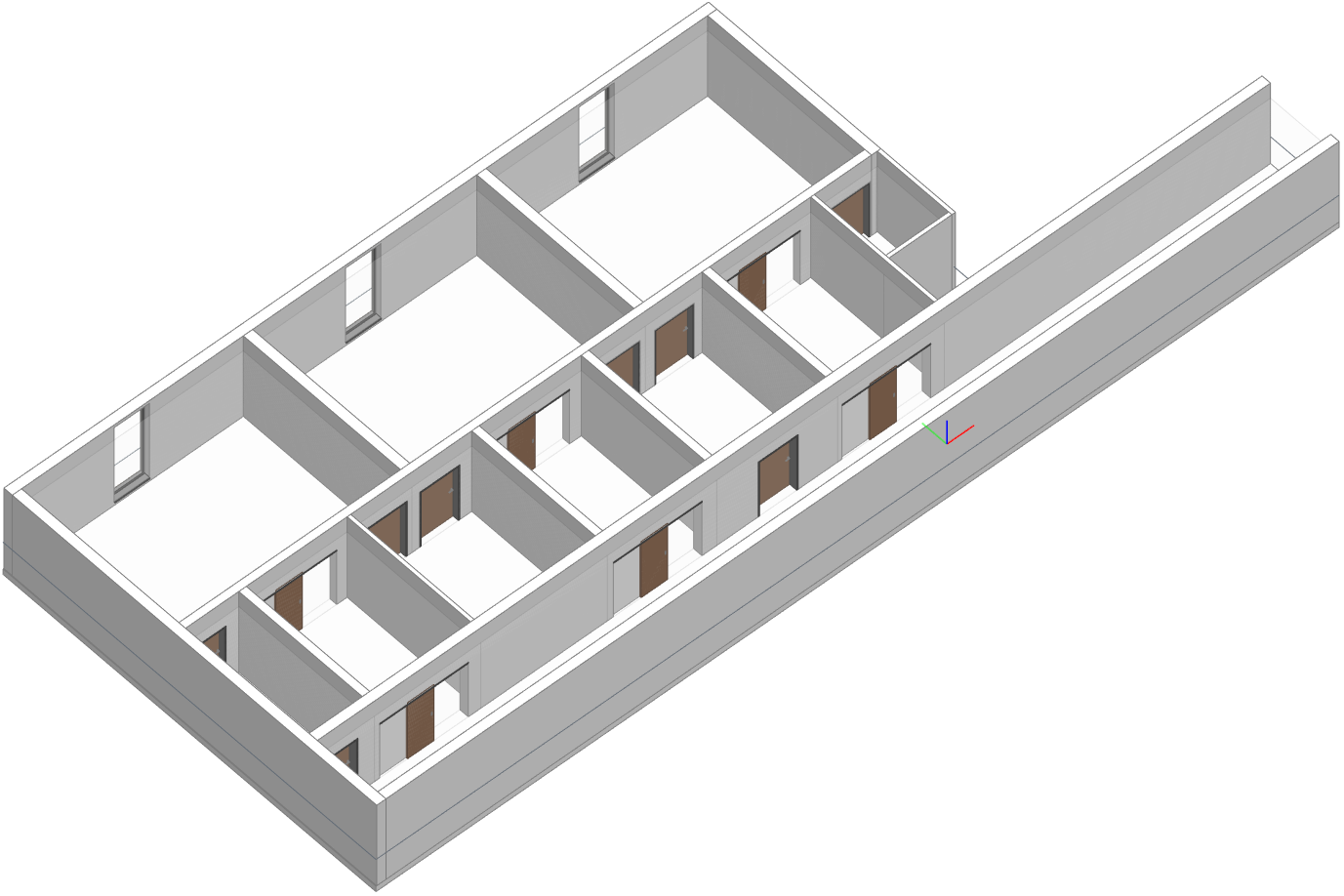} \\
    \end{tabular}
    \label{tab:DeKH_scenes}
\end{table*}

The German hospital dataset was recorded in an empty hospital with various types of facilities and rooms: offices, toilets, reception, stairs, operation rooms, and more. It boasts a variety of window and door shapes as well as some embedded furniture. For annotation, the ontology based segmentation guideline by~\cite{kaufmann2023ontology} was used. It is a construction specific joint guideline for images and point clouds.

The dataset is divided into four areas shown in~\Cref{tab:DeKH_scenes} and providing high quality semantic segmentation labels as well as manually created BIM ground truth models. The quality of the scans and labels, as well as the variety of the buildings and rooms make this dataset an important contribution to the scan-to-BIM research and the construction community.

\section{Empirical Analysis of vIoU} \label{sec:appendix_C}

\begin{figure*}[bp]
    \centering
    \begin{subfigure}[c]{0.29\textwidth}
        \centering
        \includegraphics[width=\linewidth]{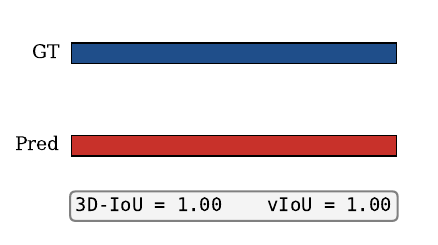}
        \caption{Identical prediction.}
        \label{fig:viou_match_identical}
    \end{subfigure}\hfill
    \begin{subfigure}[c]{0.29\textwidth}
        \centering
        \includegraphics[width=\linewidth]{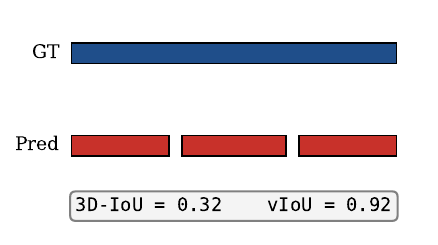}
        \caption{Fragmented prediction.}
        \label{fig:viou_match_fragmented}
    \end{subfigure}\hfill
    \begin{subfigure}[c]{0.29\textwidth}
        \centering
        \includegraphics[width=\linewidth]{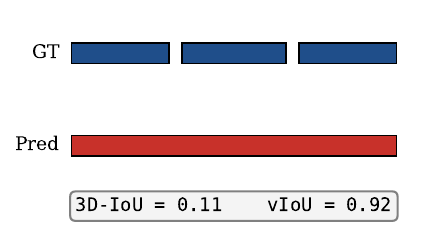}
        \caption{Merged prediction.}
        \label{fig:viou_match_merged}
    \end{subfigure}
    \vspace{0.6em}
    \caption{Response of 3D-IoU and vIoU to instance-level fragmentation and merging on a representative $5$\,m wall. Plan view; ground truth and prediction occupy the same spatial extent and are shown stacked vertically only for visual clarity. D-IoU is computed under one-to-one greedy matching averaged across ground-truth instances; vIoU is computed at the class level without matching. Unmatched predictions in (b) and unmatched ground-truth instances in (c) do not contribute to the 3D-IoU average; protocols that penalize false positives would yield even lower 3D-IoU scores in (b).}
    \label{fig:viou_matching}
\end{figure*}

To support the design choices behind vIoU, we provide an empirical characterisation of how vIoU and 3D-IoU respond to translational offsets between aligned elements, and to fragmentation or merging of the predicted geometry. For sensitivity experiments, we use an axis-aligned cuboid of dimensions $5.0 \times 0.20 \times 3.0$\,m as a representative wall element. A second copy is translated by an offset $\delta$ along a chosen axis, and we compute analytical 3D-IoU on the bounding boxes alongside vIoU using the centroid-occupancy rule with a $5$\,cm voxel size, matching the configuration used throughout the paper.

\begin{figure}[h!]
    \centering
    \begin{subfigure}[c]{\columnwidth}
        \centering
        \includegraphics[width=0.85\linewidth]{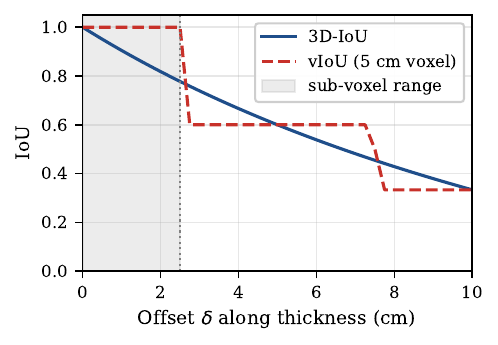}
        \caption{Sub-voxel tolerance along the thickness direction. vIoU at $5$\,cm voxel size remains exactly $1.0$ for offsets up to half a voxel ($2.5$\,cm), while 3D-IoU has dropped to $0.78$ over the same range.}
        \label{fig:viou_subvoxel}
    \end{subfigure}
    \vspace{0.2em}
    \begin{subfigure}[c]{\columnwidth}
        \centering
        \includegraphics[width=0.85\linewidth]{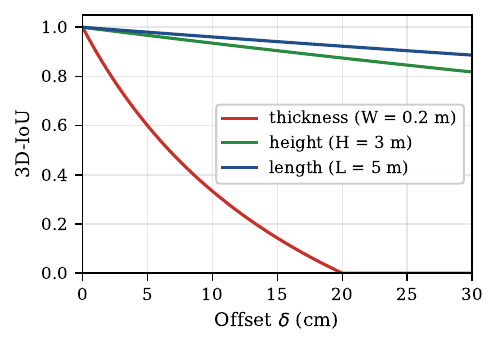}
        \caption{Per-axis sensitivity of 3D-IoU. The initial slope equals $2/L$, so the thinnest dimension dominates the response.}
        \label{fig:viou_per_axis}
    \end{subfigure}
    \vspace{0.2em}
    \caption{Sensitivity of 3D-IoU and vIoU under translational offset between two otherwise identical wall elements ($5.0 \times 0.20 \times 3.0$\,m).}
    \vspace{-1em}
    \label{fig:viou_sensitivity}
\end{figure}

\paragraph{Sub-voxel tolerance.}
\Cref{fig:viou_subvoxel} reports IoU as a function of offset along the wall thickness direction. For $\delta \leq 2.5$\,cm, that is, half the voxel size, vIoU remains at exactly $1.0$ since no voxel centroid changes its occupancy state. Over the same range, 3D-IoU monotonically decreases from $1.0$ to $0.78$. Beyond the half-voxel threshold the two metrics agree to within one quantisation step. The sub-voxel insensitivity is particularly relevant for thin elements: the fixed $2.5$\,cm tolerance corresponds to $12.5\%$ of the wall thickness in this example, while typical scan-to-BIM reconstruction errors of a few centimetres lie within the same range and would otherwise penalise an otherwise correct reconstruction.

\paragraph{Per-axis sensitivity.}
\Cref{fig:viou_per_axis} shows 3D-IoU under offsets along each principal axis individually. The initial slope at $\delta = 0$ equals $2/L$ where $L$ is the dimension along the offset direction: $0.40$/m along the length, $0.67$/m along the height, and $10$/m along the thickness. The thickness direction is therefore $25\times$ more sensitive than the length direction. This confirms that the practical sensitivity of 3D-IoU on wall-like geometries is dominated by the smallest dimension, which is the failure mode that vIoU's voxel-level tolerance is designed to address.

\paragraph{Fragmentation and merging.}
The complementary advantage of vIoU is that it operates at the class level and does not require instance-level matching between predictions and ground truth. \Cref{fig:viou_matching} illustrates this on three scenarios involving a $5$\,m wall: an identical prediction (\Cref{fig:viou_match_identical}), a fragmented prediction in which the wall is split into three correctly-placed segments (\Cref{fig:viou_match_fragmented}), and a merged prediction in which three adjacent ground-truth walls are reconstructed as a single continuous element (\Cref{fig:viou_match_merged}). The voxel coverage is nearly identical in the fragmented and merged cases, so vIoU returns the same value of $0.92$ across them. 3D-IoU under standard one-to-one greedy matching drops to $0.32$ and $0.11$ respectively, since unmatched ground-truth instances contribute zero to the average. Both fragmentation and merging are common outcomes of point-cloud-based reconstruction, making this a significant source of evaluation noise that vIoU avoids by construction.

\end{document}